\documentclass[letterpaper]{article}
\providecommand{\NewCommandCopy}[2]{\let#1#2}
\usepackage[preprint]{aaai2027}
\usepackage[hyphens]{url}
\usepackage{graphicx}
\usepackage{natbib}
\usepackage{caption}
\usepackage{amsmath}
\usepackage{booktabs}
\usepackage{array}
\newcolumntype{L}[1]{>{\raggedright\arraybackslash}p{#1}}

\title{Visual Credit Audit for Multimodal Spatial Reasoning}
\author{
Feixiang Liu\textsuperscript{\rm 1,2},
Qiang Qiu\textsuperscript{\rm 1},
Lanbo Sun\textsuperscript{\rm 1,2},
Nan Wei\textsuperscript{\rm 1},
Huawei Shen\textsuperscript{\rm 1},
Xueqi Cheng\textsuperscript{\rm 1}
}
\affiliations{
\textsuperscript{\rm 1}State Key Laboratory of AI Safety, Institute of Computing Technology, Chinese Academy of Sciences\\
\textsuperscript{\rm 2}University of Chinese Academy of Sciences\\
liufeixiang23s@mails.ucas.ac.cn, qiuqiang@ict.ac.cn, sunlanbo24s@ict.ac.cn,\\
weinan@ict.ac.cn, shenhuawei@ict.ac.cn, cxq@ict.ac.cn
}

\begin{document}

\maketitle

\begin{abstract}
Closed yes/no spatial benchmarks can reward a correct answer even when the image adds little support beyond no-image contexts. Under a fixed forced-choice interface, Visual Credit Audit (VCA) separates two estimands: whether the benchmark image gives the model's declared decision more support than text-only and blank controls, and whether the model responds to relation-specific visual evidence. The first audit is training- and label-free and does not require an answer flip. Applying labels yields dependence-credited correctness (D-CC); on correct items, it equals same-control gold-aligned positive gain, while prediction alignment extends the audit to errors. Across four open MLLMs and two spatial benchmarks, 12.73--26.25\% of decisions are correct yet uncredited. Matched same-split image permutation reduces D-CC by 21.25--47.80 points, with every paired 95\% interval above zero. Fixed-pixel relation contrasts and a $3{\times}3$ evidence-source factorial show why null controls cannot identify relation response. Among controlled correct-but-uncredited agreement decisions, response to relation reversal spans 81.57--100.00\%, while 32.11\% pooled change answer. Independently audited outcomes on 108 geometry-compatible edits provide a bounded natural-image correspondence check. VCA thereby decomposes benchmark success into correctness, additional image support, and relation-consistent response.
\end{abstract}

\noindent\textbf{Code:} \url{https://github.com/SouthWinter/VCA}

\section{Introduction}

Spatial reasoning is a basic capability for multimodal systems that must understand where objects are and how they relate. Modern MLLMs combine large-scale visual-language pretraining with strong instruction-following priors \citep{bai2023qwenvl,chen2024internvl,bai2025qwen3vl,wang2025internvl35, li2024llavaonevision}, and spatial benchmarks such as VSR and WhatsUp evaluate whether they can answer relation questions involving left/right, above/below, and other reference frames \citep{liu2023visualspatial,kamath2023whatsup, rajabi2024gsrbench}. Yet correctness on a closed yes/no question is not the same as incremental visual support. A model can answer ``yes'' to a plausible query such as ``the TV is above the cat'' because the text is familiar, the prompt format is biased, or the dataset contains strong object-relation priors. The answer is then correct even when no-image controls support the same decision at least as strongly.

This ambiguity is visible in our audit results. On GSR-COCO, Qwen reaches 90.91 original-image accuracy, but only 78.18 of all examples are both correct and more strongly supported by the image than by text-only or blank controls. For LLaVA, the corresponding values are 81.93 and 55.68. These gaps remain after removing the requirement that a no-image context must favor the opposite answer. Moreover, when the original decision direction is held fixed, a deterministic same-split image permutation reduces D-CC by 21.25--47.80 points across all eight runs, with every paired interval above zero. The core issue is therefore not simply low spatial accuracy; it is success without benchmark-image support advantage under the prespecified controls.

This distinction changes benchmark interpretation. Models with similar accuracy can differ in support-qualified performance, and aggregate success can hide whether improvements arise from image-conditioned evidence or from stronger answer priors. A decision-level audit makes these differences measurable without retraining the evaluated model. It also exposes a second distinction: no marginal image advantage does not by itself imply no visual response. When text and image redundantly support the same answer, removing the image can preserve support even though reversing its relation changes the model's score. Null controls and relation-reversed images therefore answer different identification questions.

Existing work exposes adjacent issues but does not directly assign credit to a fixed spatial decision. VQA debiasing and counterfactual evaluation reduce language priors or test consistency under changed conditions \citep{zhang2016yin,agrawal2018vqaCP,niu2021counterfactual,zhang2024what, li2024eyes}. Spatial-reasoning methods improve the underlying model with specialized data, grounded representations, 3D priors, or visual-textual reasoning \citep{chen2024spatialvlm,ma2024spatialpin,liang2025spatialvts, xu2025spatialbench}. Hallucination and contrastive-decoding methods reduce image-inconsistent outputs \citep{rohrbach2018object,li2023pope,leng2024vcd, yin2025clearsight}. These lines are complementary, but they do not ask whether a particular forced-choice spatial decision receives greater support from its image than from matched no-image contexts while remaining relation-consistent on fixed pixels.

We use \emph{visual credit} as shorthand for this operational support audit, not for internal causal credit assignment. Credit is assigned to a specified decision--context contrast, not to a latent mechanism or unconstrained generation behavior. It asks two distinct questions. First, does the original image increase support for the model's own decision relative to no-image controls? Second, does the same scoring rule distinguish the queried relation from false alternatives on fixed pixels? Conflating these conditions is misleading: a decision may be image-sensitive but relation-inconsistent, or relation-consistent yet lack a relative image-support advantage.

VCA's methodological contribution is a two-estimand identification framework, not a new name for a correct-only gain. Its support axis extends the same dual-control event label-free over every declared decision, preserving support-advantaged errors and yielding a complete accuracy--credit decomposition once labels are applied. Its relation axis addresses a different question that null inputs cannot identify: whether fixed-pixel relation separation and controlled relation reversal change the model's evidence. Matched image permutation calibrates the headline D-CC/C-U decomposition; audited edits provide bounded natural-image correspondence.

Our contributions are:
\begin{itemize}
    \item We introduce a label-free, decision-level support audit whose correct-item restriction exactly recovers same-control gold-aligned gain, while its all-decision form retains credited errors and induces a complete correctness--credit decomposition.
    \item We separate marginal image support from relation response, establish their null-control identification boundary, and add fixed-pixel semantic contrasts plus the missing relation-opposite factorial intervention.
    \item Across four MLLMs, two benchmarks, 1,800 factorial inputs per model, and 108 independently audited edits, matched calibration and controlled response tests expose distinct correct-but-uncredited behaviors.
\end{itemize}

\section{Related Work}

\subsection{Shortcut Learning and Counterfactual VQA}

Visual question answering made image-conditioned reasoning measurable at scale, but also revealed that correct answers can follow answer priors, dataset imbalance, and question-only correlations rather than visual evidence \citep{goyal2017making,zhang2016yin,agrawal2018vqaCP}. Debiasing, causal, and counterfactual VQA methods weaken language-prior branches, change the training distribution, estimate direct and indirect effects, or perturb image-question conditions to test evidence sensitivity \citep{niu2021counterfactual,zhang2024what,li2024eyes}. These works establish the central concern behind VCA: accuracy is not enough to show evidence use. VCA keeps the yes/no item fixed and asks whether its original image gives the model's spatial decision more support than matched contexts without sufficient visual evidence.

\subsection{Compositional and Spatial Relation Grounding}

Compositional vision-language evaluation tests whether models bind objects, attributes, relations, commonsense, and word order rather than matching coarse image-text statistics \citep{johnson2017clevr,thrush2022winoground}. Spatial-relation benchmarks sharpen this problem by isolating relation words, reference frames, and left/right or above/below distinctions \citep{liu2023visualspatial,kamath2023whatsup,rajabi2024gsrbench}. Recent methods improve spatial ability through specialized data, grounded representations, 3D priors, visual-textual reasoning, or explicit spatial prompting \citep{chen2024spatialvlm,ogezi2025spare,ma2024spatialpin, liang2025spatialvts,xu2025spatialbench}. These efforts primarily ask whether a model becomes more accurate or consistent on spatial tasks. VCA instead measures whether the observed success is still supported when matched probes remove or replace the relevant visual evidence, rather than by language priors or image-insensitive behavior.

\subsection{Decision-Level Visual Credit Auditing}

The closest methodological neighbors are hallucination mitigation and contrastive visual inference. Object hallucination work evaluates whether generated descriptions or yes/no probes introduce content not supported by the image \citep{rohrbach2018object,li2023pope}, and mitigation methods use robust instruction tuning, factual feedback, or contrastive decoding to reduce unsupported outputs \citep{liu2023lrv,sun2023aligning,leng2024vcd,yin2025clearsight}. Attribution and modality-use metrics ask related but distinct questions. Perceptual Score measures accuracy loss under modality permutation \citep{gat2021perceptual}; MM-SHAP aggregates unsigned token-level Shapley contributions \citep{parcalabescu2023mmshap}; and sufficiency, comprehensiveness, and FPVG evaluate retained or removed evidence \citep{deyoung2020eraser,reich2023fpvg}. Benchmark-level audits compare original and no-image accuracy: modality-importance scoring diagnoses video-QA bias \citep{park2025modality}, Fu et al. derive label-dependent per-instance modality gain from null-image controls and an aggregate Visual Dependence \citep{fu2026seeing}, phrasing-controlled tests expose textual-prior reliance \citep{singla2026see}, and Visual Dependency Index measures aggregate loss under question-agnostic image descriptions \citep{chen2026omnimapbench}. Korkut et al. combine confidence-weighted blind correctness and gold-answer loss reduction into a continuous per-question Shortcut Score for benchmark filtering \citep{korkut2026accuracy}. VisNec uses gold-response loss reduction under blind visual-token attention for data selection \citep{dong2026visnec}. Visual-degradation audits expose support weakening or invariant answers \citep{lan2026seeingwithout}, while ViewDiag tests spatial evidence insensitivity across viewpoints \citep{bhat2026consistent}. Multimodal interaction decompositions explicitly model redundancy and synergy \citep{yu2024mmoe}, while conflicting-text evaluations test whether vision can override textual misinformation \citep{zhang2026images}. These results motivate separating marginal credit from counterfactual source response rather than treating either as a proxy for the other.

On correct items, VCA's dependence event intentionally reduces exactly to a gold-aligned positive gain under the same controls; the contribution is not a different correct-only formula. VCA instead completes this event without labels over every declared decision, retaining support-advantaged errors, and places the resulting support decomposition beside a separately identified relation-response axis. Aggregate drops cannot assign an event to each decision, gold-only gains cannot audit errors, and flip-only rules conflate reduced support with evidence for the opposite answer. Fixed multi-verbalizer and pair-centered checks test transport across answer surfaces rather than assuming it. This relative support audit is complementary to probabilistic calibration, which measures confidence reliability \citep{naeini2015obtaining,guo2017calibration}.

\begin{figure*}[t]
\centering
\includegraphics[width=\textwidth]{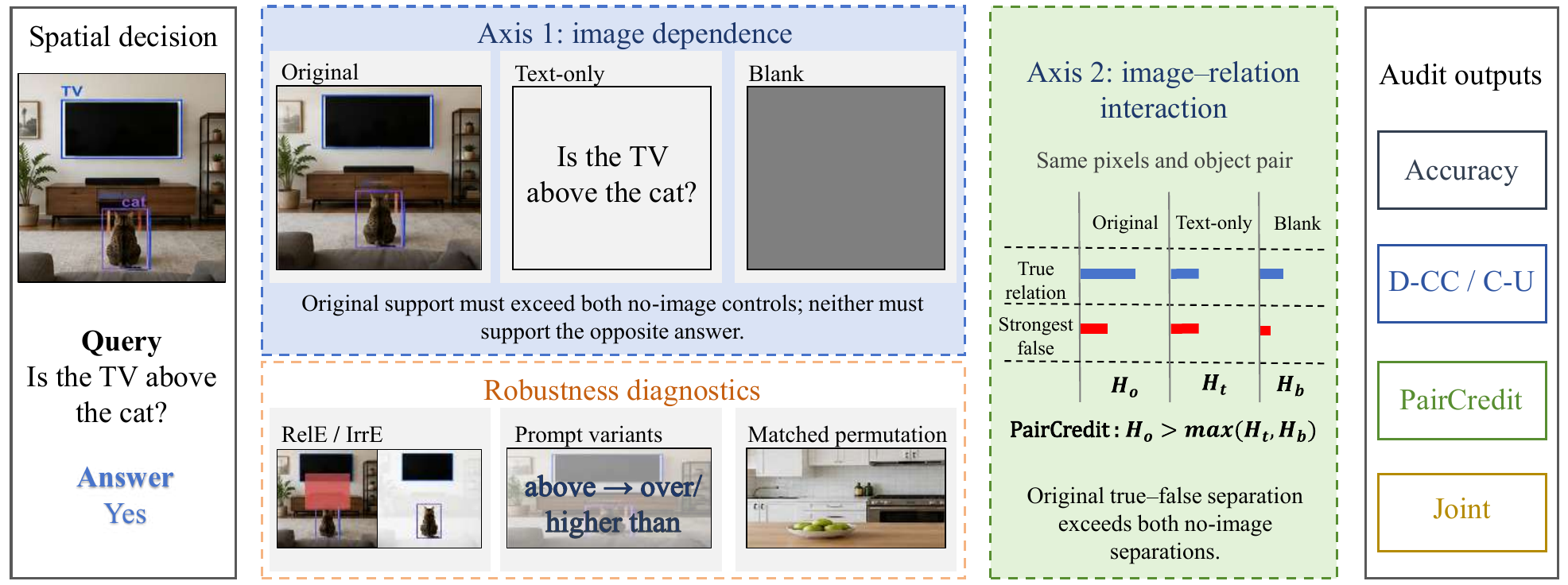}
\caption{VCA separates relative image support from image--relation interaction. The support audit compares original-image support with text-only and blank controls without requiring either control to favor the opposite answer. Fixed-image SRC contrasts true and false relations; PairCredit tests whether the original separation exceeds both no-image separations, while Joint adds absolute relation consistency and paired dependence. Matched permutation calibrates D-CC/C-U; erasures and prompt variants are diagnostics.}
\label{fig:method_overview}
\end{figure*}

\section{Method}

\subsection{Problem Setup and Two-Axis Audit}

Each sample is a binary spatial query $(I,q,y)$, where $I$ is an image, $q$ asks about a spatial relation, and $y\in\{\mathrm{yes},\mathrm{no}\}$ is used only for offline correctness. VCA fixes a binary answer interface, freezes the MLLM, and caches literal yes/no continuation margins under matched contexts. It separates two conditions that a spatial answer should satisfy. \emph{Relative image dependence} asks whether the original image provides more support for that interface decision than no-image controls. \emph{Relation consistency} asks whether the same image and object pair support the true relation over false alternatives. The first condition is prediction-aligned and benchmark-label-free; the second is an offline semantic check using the benchmark's paired relation statements. Figure~\ref{fig:method_overview} summarizes this separation. Here ``credit'' denotes an operational, decision-level within-model audit; pixel-level causal attribution lies outside this protocol.

\subsection{Prediction-Aligned Image Support}

Let $\ell_\theta(a\mid c,q)$ be the frozen MLLM's negative log-likelihood for answer $a$ under context $c$. The yes margin $m_c=\ell_\theta(\mathrm{no}\mid c,q)-\ell_\theta(\mathrm{yes}\mid c,q)$ is positive when the model prefers ``yes.'' We record its original prediction direction $d_i=+1$ if $m_{i,o}\geq0$ and $d_i=-1$ otherwise, and define prediction-aligned support $r_{i,c}=d_i m_{i,c}$. Thus larger $r_{i,c}$ means stronger support for the model's original decision, whether or not that decision is correct.

The primary no-image controls are text-only ($t$) and blank ($b$). We define
\begin{equation}
G_i^{\mathrm{img}}=r_{i,o}-\max(r_{i,t},r_{i,b}),\qquad
D_i={\bf 1}\{G_i^{\mathrm{img}}>0\}.
\end{equation}
We call $G_i^{\mathrm{img}}$ the VCA dependence gap. The zero threshold marks the estimand boundary for additional support over both controls; it is neither a significance test nor a fitted calibration cutoff. Throughout, ``dependence'' denotes this control-relative support advantage, not answer-level necessity. Crucially, it does not require missing evidence to make the model choose the opposite binary answer. After the audit, the benchmark label gives $\kappa_i={\bf 1}\{\hat y_i=y_i\}$ and the dependence-credited correct rate
\begin{equation}
\mathrm{D\mbox{-}CC}=N^{-1}\sum_i D_i\kappa_i.
\end{equation}
Because \(D_i\) is assigned before labels are consulted, it induces a complete four-cell decomposition:
\begin{equation}
\begin{aligned}
\mathrm{Acc}&=\mathrm{E}[\kappa D]+\mathrm{E}[\kappa(1-D)],\\
\mathrm{Dep}&=\mathrm{E}[\kappa D]+\mathrm{E}[(1-\kappa)D],\\
1&=\mathrm{E}[\kappa D]+\mathrm{E}[\kappa(1-D)]\\
&\quad+\mathrm{E}[(1-\kappa)D]+\mathrm{E}[(1-\kappa)(1-D)].
\end{aligned}
\label{eq:four_cell}
\end{equation}
The first cell is D-CC and the second is correct-but-uncredited (C-U). The third retains support-advantaged errors, which a correct-only gain omits; the fourth contains errors without control-relative image advantage. Thus one prespecified event audits both sides of correctness rather than changing its meaning after observing the label. We also report credited precision among $D_i=1$ decisions. A stricter isolation diagnostic additionally requires $r_{i,t}<0$ and $r_{i,b}<0$; it is retained for sensitivity analysis, not used as the primary definition. Separately, decision survival records whether the original binary answer persists under each no-image control; this distinguishes lost support from answer-level necessity.

\paragraph{Properties of the raw audit.} Within a fixed binary answer interface, prediction alignment is formally sign-symmetric: exchanging the names of its two answer tokens flips both the margin and $d_i$, leaving every $r_{i,c}$ and $G_i^{\mathrm{img}}$ unchanged. This algebraic property does not assert invariance to a different verbalizer; we test that separately with full True/False scoring. The audit therefore asks a relative question about original-image support, rather than whether ``yes'' or ``no'' has a larger absolute logit. Finally, $D_i$ is deliberately independent of the benchmark label. This separates three outcomes that accuracy alone conflates: a correct support-advantaged answer, a correct answer no better supported by the image than by no-image controls, and a support-advantaged error. Let \(s_i\in\{-1,+1\}\) denote the gold-answer direction. Because \(\kappa_i=1\) implies \(d_i=s_i\), the correct-item restriction is exactly
\begin{equation}
\kappa_iD_i=\kappa_i{\bf 1}\!\left\{
s_i m_{i,o}>\max(s_i m_{i,t},s_i m_{i,b})\right\}.
\end{equation}
Thus prediction alignment does not manufacture a new correct-only score. It extends the same order event label-free to all decisions, where support-advantaged errors remain visible rather than disappearing from a gold-only statistic. Accordingly, Dep is a prespecified within-item order event, not a probability of image use. Its matched calibration holds \(d_i\) fixed and replaces only the original image by a deterministic same-split permutation, directly testing whether a generic visual input reproduces the observed support advantage. Raw dependence gaps are interpreted within model because likelihood scales need not define a shared cross-model ability scale.

\paragraph{Identification boundary of null controls.} For fixed query text \(q\), observations at a relation-supporting image \(I^+\), text-only input \(\emptyset\), and blank image \(B\) do not constrain the response at an unobserved relation-opposite image \(I^-\). Without assumptions linking these contexts, two frozen response functions can agree on \(\{m(I^+,q),m(\emptyset,q),m(B,q)\}\) yet place \(m(I^-,q)\) on opposite sides of \(m(I^+,q)\). Therefore no null-control statistic, including \(G^{\rm img}\), identifies the sign of \(\Delta_V(q)=m(I^+,q)-m(I^-,q)\). This boundary separates VCA's marginal support estimand from the relation responsiveness tested by the controlled factorial below.

\paragraph{Answer-interface protocol.} The canonical Yes/No interface is prespecified and covers every item, so it supplies the complete dataset-level report without outcome-dependent interface selection. It is not a uniquely privileged semantic surface. The fixed Yes/No--True/False mean-margin ensemble is an aggregate sensitivity analysis, not a replacement definition. Where inverse relation statements exist, pair-centered semantic differencing removes offsets shared by the true and false queries and supplies the relation-specific interface check.

\subsection{Contexts and Diagnostic Interventions}

Text-only omits the image, while blank preserves the multimodal prompt format with a same-size fixed mid-gray canvas. These two no-image controls form the primary dependence audit. A deterministic same-split mismatch is useful for testing whether the signal merely detects a normal image rather than queried evidence. We run this natural unrelated-image check on all eight model--dataset combinations. Because object absence can itself make a binary spatial statement false, mismatch is reported separately and does not determine primary credit. Relation-erased (RelE) and irrelevant-erased (IrrE) views are artifact-sensitive box-erasure diagnostics; SRC supplies the pixel-preserving relation test. RelE removes the padded subject--object union, while IrrE checks preservation under nuisance removal. Figure~\ref{fig:method_overview} summarizes these roles; exact constructions are given in the supplement.

\subsection{Relation Consistency, Joint Audit, and Derived Interaction}

For an image $I$, object pair $(u,v)$, true relation $r^+$, and false alternatives $\mathcal{R}^-$, Semantic Relation Contrast (SRC) scores all statements without changing the image. A group is relation-consistent when the true statement is accepted and every false alternative is rejected. In each context $c\in\{o,t,b\}$, define the true--false separation and its image interaction as
\begin{equation}
\begin{aligned}
H_c&=m_c(q_{r^+})-\max_{r^-}m_c(q_{r^-}),\\
G^{\mathrm{int}}&=H_o-\max(H_t,H_b).
\end{aligned}
\end{equation}
These contrasts define \(\mathrm{PairCorrect}={\bf 1}\{H_o>0\}\) and \(\mathrm{PairCredit}={\bf 1}\{G^{\mathrm{int}}>0\}\). Additive answer-surface offsets shared by the true and false queries cancel in \(H_c\). PairCorrect ranks the pair, whereas strict \(\mathrm{Rel}_o\) requires absolute acceptance and rejection. Let Pair D require $D_i=1$ for the true statement and every false alternative. VCA Joint is their transparent conjunction:
\begin{equation}
\mathrm{Joint}\equiv\mathrm{Rel}_o\wedge\mathrm{PairD},\qquad
\mathrm{Joint}=1\Longrightarrow G^{\mathrm{int}}>0.
\label{eq:joint_implies_interaction}
\end{equation}
The conjunction structurally excludes nonpositive interaction; its short proof is supplementary. We report $G^{\mathrm{int}}$ over all groups as the empirical interaction diagnostic beyond the Joint subset. SRC and Joint use no edited pixels or learned scorer.

\section{Experiments}

\begin{figure*}[t]
\centering
{\small
\setlength{\tabcolsep}{0.8pt}
\begin{tabular*}{\textwidth}{@{\extracolsep{\fill}}llrrrrr@{}}
\toprule
Model & Data & Acc & Dep [CI] & D-CC [CI] & C-U &
\(\Delta_{\rm DCC}^{\rm perm}\) [CI] \\
\midrule
Qwen & VSR & 79.42 & 71.10 [68.61,73.50] & 58.53 [55.72,61.25] & 20.90 & 27.30 [23.38,31.31] \\
Qwen & GSR-COCO & 90.91 & 81.82 [79.20,84.32] & 78.18 [75.45,80.91] & 12.73 & 38.86 [35.23,42.50] \\
InternVL & VSR & 58.69 & 57.01 [54.28,59.81] & 39.55 [36.91,42.35] & 19.14 & 26.02 [23.46,28.74] \\
InternVL & GSR-COCO & 58.41 & 42.84 [39.66,46.14] & 34.43 [31.25,37.61] & 23.98 & 21.25 [18.64,23.86] \\
LLaVA & VSR & 69.90 & 76.38 [73.90,78.70] & 51.56 [48.68,54.36] & 18.33 & 47.80 [44.92,50.76] \\
LLaVA & GSR-COCO & 81.93 & 64.55 [61.14,67.73] & 55.68 [52.39,58.98] & 26.25 & 42.61 [39.77,45.34] \\
Ministral & VSR & 64.45 & 63.33 [60.61,65.89] & 48.52 [45.80,51.24] & 15.93 & 34.43 [31.63,37.23] \\
Ministral & GSR-COCO & 71.48 & 59.77 [56.48,62.95] & 55.34 [52.05,58.75] & 16.14 & 33.30 [30.57,36.02] \\
\bottomrule
\end{tabular*}}
\captionof{table}{Canonical Yes/No relative-support results. D-CC additionally requires correctness; C-U is Acc\(-\)D-CC. \(\Delta_{\rm DCC}^{\rm perm}\) is original D-CC minus D-CC after a prespecified same-split image permutation, retaining the original decision and correctness label. CIs are 95\% paired-bootstrap intervals from 5,000 resamples.}
\label{tab:main_results}
\vspace{3pt}
\includegraphics[width=0.94\textwidth]{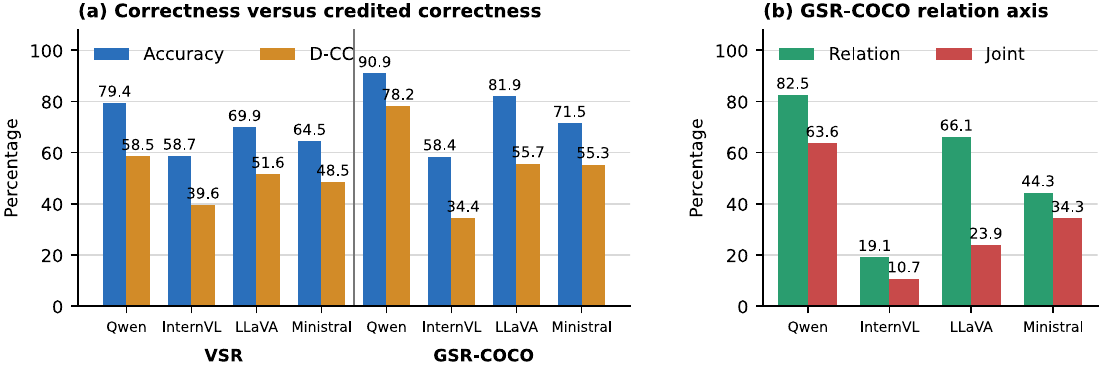}
\captionof{figure}{VCA decomposes correctness and control-relative support; same-image relation consistency and paired dependence combine into Joint.}
\label{fig:results_overview}
\end{figure*}

\subsection{Experimental Setup}

\paragraph{Benchmarks and backbones.}

VSR \citep{liu2023visualspatial} is the primary benchmark because it has broad spatial-relation coverage and boxes for optional stress diagnostics. Its test split contributes 1,249 queries spanning nine relation types. We also evaluate GSR-COCO, a spatial yes/no split derived from COCO \citep{lin2014coco}: 880 statements form 440 same-image true/false relation pairs over left/right and above/below. We evaluate four open 7B--14B autoregressive families: Qwen3-VL-8B-Instruct \citep{bai2025qwen3vl}, InternVL3.5-8B \citep{wang2025internvl35}, LLaVA-OneVision-Qwen2-7B-OV \citep{li2024llavaonevision}, and Ministral-3-14B-Instruct-2512 \citep{mistralai2026ministral3}, abbreviated as Qwen, InternVL, LLaVA, and Ministral. The primary audit requires no boxes; mismatch and box-based views remain supplementary diagnostics.

\paragraph{Metrics and protocol.}

The primary support report contains Accuracy, the label-free dependence event rate (Dep), D-CC, and C-U \(=\) Acc\(-\)D-CC. Dep is the event rate used to audit all decisions, not a separate ability score. On GSR-COCO, paired true/false statements add fixed-image relation consistency, PairCredit, and their stricter Joint audit; VSR supports the general image-support estimand but not this full relation axis. Canonical Yes/No is prespecified for complete rates. A fixed Yes/No--True/False ensemble and pair-centered semantic contrasts test answer surface transport. All headline values use raw margins from the frozen MLLM; no scorer is trained or selected, and the threshold $G^{\mathrm{img}}>0$ is fixed before labels are consulted. Primary comparisons fix the query and rule while changing only the visual input. Strict isolation, mismatch, erasures, and direction splits are supplementary. Confidence intervals use 5,000 paired-bootstrap resamples; table values are percentages unless noted. Canonical Acc/Dep/D-CC/C-U, matched D-CC image-permutation excess, and controlled factorial contrasts form the headline report; other controls test robustness and edited outcomes test external correspondence. The submission materials include source code, exact prompt serialization, per-context losses and margins, deterministic data builders, factorial inputs, and deidentified edit audits.

\subsection{Main Results}

\paragraph{Image-support advantage.} Table~\ref{tab:main_results} shows that 12.73--26.25\% of all decisions are correct but uncredited, with substantial variation across models and datasets. Dep can exceed accuracy because a support-advantaged decision may still be wrong; D-CC requires both properties. The all-decision extension retains 3.64--24.82\% support-advantaged errors across the eight runs, a cell absent from correct-only gain. Under the matched image permutation, original-image D-CC is 21.25--47.80 points higher in every run, and all paired intervals exclude zero. Equivalently, C-U increases by the same amount because accuracy and correctness labels remain fixed.

\paragraph{Relation consistency and joint credit.} The second axis resolves a distinct ambiguity: relation consistency and paired image dependence capture complementary properties. Their conjunction defines Joint, while all-pair $G^{\rm int}$ is the empirical interaction diagnostic.

\par\noindent\begin{minipage}{\columnwidth}
\centering
{\small
\setlength{\tabcolsep}{1.2pt}
\begin{tabular*}{\columnwidth}{@{\extracolsep{\fill}}lrrr@{}}
\toprule
Model & Rel.$_o$ & PairCredit & Joint \\
\midrule
Qwen & 82.50 & 92.95 & 63.64 \\
InternVL & 19.09 & 62.05 & 10.68 \\
LLaVA & 66.14 & 88.86 & 23.86 \\
Ministral & 44.32 & 85.45 & 34.32 \\
\bottomrule
\end{tabular*}}
\captionof{table}{GSR-COCO relation axis over 440 pairs. Rel.$_o$ is strict original-image consistency; PairCredit tests image-specific true--false separation; Joint additionally requires paired dependence.}
\label{tab:relation_decomposition}
\end{minipage}\par\smallskip

PairCredit holds for 62.05--92.95\% of pairs, showing image-specific semantic separation beyond the stricter Joint subset. Joint retains only pairs that also satisfy absolute relation consistency and paired dependence.

\paragraph{Controlled evidence-source factorial.} To identify the response hidden by null controls, we cross visual and textual evidence states \(\{+,-,\emptyset\}\) over 200 balanced spatial stems and four directions. This gives nine inputs per stem (1,800 per model): each source supports the query, supports its opposite, or supplies no relation. For a left-of query, for example, the image places the subject left, right, or axis-neutral, while an evidence note states left, right, or only names the objects. Within a textual state \(s_T\), the controlled visual response is
\begin{equation}
\Delta_V(s_T)=m(I^+,s_T)-m(I^-,s_T).
\end{equation}
A positive value is a correct-direction response to changing only the rendered relation. We focus on correct agreement cells \((I^+,s_T^+)\) and \((I^-,s_T^-)\), where both sources support the same answer. If \(G^{\rm img}\leq0\) but \(\Delta_V(s_T)>0\), the decision is marginally uncredited yet counterfactually relation-responsive; an answer change under the same intervention is the stricter flip criterion. To separate relation evidence from object presence, each agreement image is also paired with a same-object, same-background control aligned on the queried axis: \(N^+=d[m(I^{\rm rel},s_T)-m(I^{\rm neutral},s_T)]>0\).

\par\noindent\begin{minipage}{\columnwidth}
\centering
{\small
\setlength{\tabcolsep}{1.0pt}
\begin{tabular*}{\columnwidth}{@{\extracolsep{\fill}}lrrrrrr@{}}
\toprule
Model & \shortstack{Correct\\\(n\)} & \shortstack{C-U\\\(n\)} &
\shortstack{C-U\\\%} & Resp. & \(N^+\) & Flip \\
\midrule
Qwen & 400 & 305 & 76.25 & 100.00 & 100.00 & 34.43 \\
InternVL & 400 & 331 & 82.75 & 81.57 & 96.07 & 21.75 \\
LLaVA & 310 & 8 & 2.58 & 100.00 & 75.00 & 100.00 \\
Ministral & 400 & 38 & 9.50 & 100.00 & 100.00 & 89.47 \\
Pooled & 1,510 & 682 & 45.17 & 91.06 & 97.80 & 32.11 \\
\bottomrule
\end{tabular*}}
\captionof{table}{Evidence-source factorial. Correct \(n\) is out of 400 agreement decisions per model; C-U \(n\) and rate are conditioned on correct decisions, and Resp./\(N^+\)/Flip on C-U.}
\label{tab:factorial_main}
\end{minipage}\par\smallskip

Table~\ref{tab:factorial_main} resolves the apparent ambiguity rather than erasing it. Of 1,510 correct agreement decisions, 682 (45.17\% [43.62,46.75]) lack marginal VCA credit; 91.06\% [89.21,92.83] of that subset moves in the correct direction under relation reversal, and 32.11\% [29.34,34.81] changes its binary answer. Qwen and InternVL contribute 636 of the 682 C-U decisions; their responsiveness is 100.00\% [98.76,100.00] and 81.57\% [77.04,85.38], respectively. Moreover, 97.80\% [96.71,98.82] exceed the object-preserving neutral control (Qwen 100.00\%; InternVL 96.07\% [93.87,97.94]).

\paragraph{Human-audited edit set.} Of 440 seeded groups, 114 were geometry-feasible before model scoring. Three consenting non-author volunteers independently assessed shuffled, blinded batches with repeat checks; institutional policy required no formal review for this low-risk task. Consensus retained 108 edits. The same accepted set evaluates all four models. Feasibility favors smaller boxes with sufficient destination clearance, so this check applies to geometry-compatible edits rather than the full benchmark.

\paragraph{Independent edited-outcome check.} SRC establishes strict original-image relation consistency (Rel.$_o$) before editing. Table~\ref{tab:relation_validity_main} adds an independent edit outcome: the frozen model must recognize the accepted flip with both objects visible.

\par\noindent\begin{minipage}{\columnwidth}
{\small
\setlength{\tabcolsep}{0.8pt}
\begin{tabular*}{\columnwidth}{@{\extracolsep{\fill}}lrrrrr@{}}
\toprule
Model & R/J \(n\) & Rel. & +Vis. & Rel.$_o$ & Joint \\
\midrule
Qwen & 95/79 & 84.26 & 67.59 & 71.58 & 78.48 \\
InternVL & 25/15 & 25.93 & 25.93 & 36.00 & 33.33 \\
LLaVA & 80/30 & 62.96 & 57.41 & 70.00 & 83.33 \\
Ministral & 58/46 & 45.37 & 41.67 & 50.00 & 56.52 \\
Macro & 258/170 & 54.63 & 48.15 & 56.89 & 62.92 \\
\bottomrule
\end{tabular*}}
\captionof{table}{Edited-outcome check on 108 accepted geometry-compatible edits. R/J \(n\) gives Rel.$_o$/Joint counts per model. Rel. tests the edited true/false pair after human visibility audit; +Vis. additionally requires the model's two object-presence responses. Rel.$_o$/Joint use +Vis.}
\label{tab:relation_validity_main}
\end{minipage}\par\smallskip

On the accepted swaps, edited relation-pair consistency is 54.63\% overall and 69.47\% within Joint; adding model object-presence responses yields 48.15\% overall, 56.89\% within Rel.$_o$, and 62.92\% within Joint. Among the 100 pooled model--edit records that are source-correct but uncredited, 25.00\% [17.20,33.05] still achieve flip+visibility. This natural correspondence does not make patch swaps a clean intervention, but it confirms that responsive uncredited cases are not confined to procedural renders. The comparable threshold-crossing rates are therefore 32.11\% for controlled answer flip and 25.00\% for natural flip+visibility; 91.06\% is the weaker directional-margin event.

\subsection{Robustness Analysis}

\paragraph{Matched image calibration.} The primary calibration preserves the audited decision: it fixes the original direction and replaces only the image by a deterministic same-split permutation. Original-image D-CC exceeds matched-permutation D-CC by 21.25--47.80 points, with every paired 95\% interval above zero (Table~\ref{tab:main_results}). At item level, correct original-only assignments occupy 22.73--49.72\% of each run, versus 0.91--15.13\% for permutation-only assignments. This tests whether unrelated visual content reproduces headline credit, not merely the all-decision Dep rate. Alternate control aggregations and visual controls are supplementary.

\paragraph{Answer-surface robustness.} The fixed two-interface ensemble has 72.39--97.44\% D-CC agreement with the canonical report. Pair-centered semantic contrast further removes offsets shared by true and false relation queries; across 440 GSR groups and 504 clean VSR inverse pairs, Yes/No--True/False PairCredit Jaccard is 62.80--97.60\%, improving agreement in all eight comparisons. These are transport checks, not claims that the canonical surface is itemwise invariant.

\paragraph{Threshold and probe robustness.} Only 0.08--0.72\% of items per run have \(|G^{\rm img}|\leq0.01\). Across fixed-blank, mean-color, and mismatch-inclusive control sets, 89.78\% of pooled assignments agree across all three, and 98.09\% of headline correct-but-uncredited decisions remain in their intersection. Full run-level stability and continuous agreement are supplementary.

\paragraph{Baseline utility and complementarity.} At 50\% leave-one-model-out coverage, retained original-minus-text-only accuracy (``Visual Gain'') is 33.64 for VCA, comparable to Shortcut Score's 33.92 \citep{korkut2026accuracy} and confidence's 32.90, and above VisNec-style's 30.52 ($+3.12$ [$+1.32$,$+4.91$]). VCA overlaps 64.74\% with confidence at this coverage. These results support complementary audit and filtering use, not universal predictive dominance.

\section{Discussion}

VCA yields a layered report rather than one replacement for accuracy. Accuracy retains task success; D-CC identifies successes with additional image support; PairCredit tests image-specific semantic separation, while Rel.$_o$ and Joint retain stricter absolute checks when paired statements exist. Raw gaps support ranking, while edited outcomes check semantic transfer. Together these outputs preserve the distinction between correctness, relative support, and relation discrimination.

The four-cell decomposition in Eq.~\ref{eq:four_cell} also gives two distinct diagnoses that the same accuracy cannot. C-U mass marks decisions whose correctness outruns support from the benchmark image under the chosen controls. Support-advantaged error mass instead marks decisions for which image-specific evidence is present but mapped to the wrong answer. A gold-only gain necessarily collapses this second distinction because it cannot credit evidence for an incorrect declared decision. Prediction alignment preserves it without altering the correct-item event.

Operationally, Accuracy and D-CC compare task success with support-qualified success; the off-diagonal cells distinguish unsupported correctness from support-advantaged error; and the relation axis asks whether evidence identifies the queried relation. These levels share cached margins but retain separate interpretations. VCA is a post-hoc companion to accuracy, not a fitted replacement score.

Text-only and blank define the marginal support contrast; alternate controls test its stability. The factorial supplies the relation-opposite intervention that null controls cannot identify, separating responsive and nonresponsive uncredited decisions. Its neutral control distinguishes relation evidence from generic object presence, while audited swaps connect the distinction to independently judged natural-image edits.

Crucially, the factorial does not replace VCA as the headline audit. It requires relation-opposite images that most natural benchmarks cannot supply, whereas VCA uses the benchmark input and source-agnostic null controls without labels. Their roles are complementary: VCA scales to quantify marginal credit on existing tests; the factorial identifies why an uncredited decision can still respond to vision. These two estimands should not be forced into one quantity.

Two design choices make this decomposition auditable. Prediction alignment anchors the model's declared decision, retaining support-advantaged errors instead of folding them into a gold-answer score. The maximum over text-only and blank requires the image to exceed each control. Alternate blanks, mismatch, and edits test implementation, normal-image sufficiency, and semantic transfer without redefining the headline event. Fixed zero-threshold events coexist with continuous ranking, and matched image calibration establishes that the aggregate advantage is specific to the benchmark image.

\section{Limitations}

VCA measures relative support under prespecified controls, not an internal causal pathway or pixel attribution. Its per-question rates are defined for a prespecified forced-choice interface; fixed multi-verbalizer and pair-centered analyses evaluate transport across equivalent answer surfaces. The factorial identifies behavioral response to controlled relation changes in simple procedural scenes, not an internal causal path; the natural-benchmark edit check covers a geometry-selected subset. The evaluation target is a fixed forced-choice decision interface across two spatial benchmarks and four autoregressive MLLMs, not unconstrained greedy text. Open-ended generation and other reasoning domains require output-specific support definitions.

\section{Conclusion}

VCA provides a training-free audit of whether spatial decisions gain support from the image and distinguish relations on fixed pixels. Across four MLLMs and two benchmarks, it quantifies 12.73--26.25\% correct-but-uncredited mass hidden by aggregate accuracy; matched image permutations confirm 21.25--47.80 points of original-image D-CC excess, with all paired intervals positive. An identification boundary and controlled evidence-source factorial show why this marginal-credit event must be separated from relation responsiveness: within-model directional response to visual reversal spans 81.57--100.00\%, and 97.80\% pooled exceed the object-preserving neutral control. Pair-centered contrasts improve answer-surface stability; audited edits check natural correspondence; cached margins make all outputs reproducible without retraining.

\bibliography{vca_aaai2027}

\clearpage
\section{Supplementary Material}

\section{Audit Definitions and Protocol}

\subsection{Prediction-Aligned Relative Image Dependence}

For every binary query, we cache the literal yes/no continuation margin
\begin{equation}
m_{i,c}=\ell_\theta(\mathrm{no}\mid c,q_i)
       -\ell_\theta(\mathrm{yes}\mid c,q_i)
\end{equation}
for the original image \(o\), text-only input \(t\), and blank image \(b\). The original-image prediction determines \(d_i\in\{-1,+1\}\), and \(r_{i,c}=d_i m_{i,c}\) expresses support for that prediction under context \(c\). The primary continuous signal and binary audit event are
\begin{equation}
G_i^{\mathrm{img}}=r_{i,o}-\max(r_{i,t},r_{i,b}),\qquad
D_i=\mathbf{1}\{G_i^{\mathrm{img}}>0\}.
\end{equation}
Here dependence denotes a control-relative support advantage, not the necessity of the image for preserving the binary answer. No benchmark label, learned scorer, or development threshold enters \(D_i\). Labels are used only afterward for original accuracy and \(\mathrm{D\mbox{-}CC}=N^{-1}\sum_iD_i\mathbf{1}\{\hat y_i=y_i\}\). For a correct prediction, \(d_i\) equals the gold-answer direction. Thus its \(D_i\) indicator is algebraically identical to thresholding a gold-aligned per-instance gain built from the same controls. Prediction alignment contributes a label-free event over all decisions, including errors; it does not create a different correct-only indicator.

Text-only removes the image channel. Blank uses a same-size fixed mid-gray image and therefore controls for the multimodal prompt format. A random same-split image is retained as a mismatch robustness probe, but not as a no-evidence premise: object absence can legitimately make a binary statement false. Relation-erased and irrelevant-erased views are similarly optional box-erasure diagnostics. They do not determine \(G_i^{\mathrm{img}}\), \(D_i\), or D-CC.

\subsection{Relation Consistency, Joint Audit, and Derived Interaction}

GSR-COCO contains 440 object-pair groups, each with a true and false relation statement on the same image. Relation consistency requires accepting the true statement and rejecting its paired false statement. For context \(c\in\{o,t,b\}\), let \(H_c=m_c(q_{r^+})-\max_{r^-}m_c(q_{r^-})\) and \(G^{\rm int}=H_o-\max(H_t,H_b)\). Pair dependence requires both statements to satisfy \(D_i=1\), and VCA Joint is exactly \(\mathrm{Rel}_o\wedge\mathrm{PairD}\). Under \(\mathrm{Rel}_o\), the true statement's original prediction is Yes and the false statement's is No. Pair D therefore gives \(m_o^+>m_c^+\) and \(m_o^-<m_c^-\) for each \(c\in\{t,b\}\), implying \(H_o>H_c\) and hence \(G^{\rm int}>0\). The implication is a structural property of Joint. Interaction over all groups and edited outcomes provide the empirical checks beyond that property. This construction edits no pixels and trains no semantic classifier.

\subsection{Evaluation Protocol}

VSR contains 1,249 queries. GSR-COCO contains 880 statements arranged into 440 true/false pairs. Prespecified canonical Yes/No scoring supplies the complete benchmark-level headline layer. We evaluate frozen Qwen3-VL-8B-Instruct, InternVL3.5-8B, LLaVA-OneVision-Qwen2-7B-OV, and Ministral-3-14B-Instruct-2512 with identical literal yes/no forced-choice scoring: lower continuation loss determines the binary decision. This is the declared evaluation interface rather than unconstrained generation. Primary audit comparisons fix the question and decision rule while changing only the visual input. Confidence intervals for Dep and D-CC use 5,000 paired bootstrap resamples with seed 20260719. Across the 8,516 main records, no original margin is exactly zero or within \(10^{-12}\) at stored precision; the implementation maps an exact equality to Yes.

Canonical Yes/No supplies complete dataset-level rates, while a fixed Yes/No--True/False mean-margin ensemble summarizes aggregate surface robustness. Where inverse relations exist, pair-centered semantic differencing removes offsets shared by the true and false queries. Canonical Acc/Dep/D-CC, matched image-permutation excess, and the controlled factorial contrasts form the headline report; control variants test robustness, while edited-outcome regressions and filtering evaluate external correspondence.

\section{Controlled Evidence-Source Factorial}

\subsection{Identification and Construction}

For fixed textual evidence state \(s_T\), original, text-only, and blank observations constrain a frozen response function only at \(I^+\), \(\emptyset\), and \(B\). A second response function can take identical values at these three inputs and an arbitrary different value at a relation-opposite \(I^-\). Therefore no null-control statistic alone identifies whether a decision with \(G^{\rm img}\leq0\) is visually unresponsive or responds to visual evidence that is redundant with text. This is an observational non-identifiability statement; it does not depend on the use of a maximum rather than a mean or on the zero threshold.

We supply the missing intervention with a \(3{\times}3\) visual--textual factorial. Each source independently takes one of three states: supports the queried relation (\(+\)), supports its opposite (\(-\)), or supplies no relation (\(\emptyset\)). We render 50 stems for each of left, right, above, and below, giving 200 factorial groups and 1,800 inputs per model. The left-of cells, for example, place the subject left, right, or aligned on the horizontal axis, while the matched evidence note states left, right, or only the two object identities. The 512-by-512 scenes use deterministic object identities, layouts, and seed 20260724. Across the three visual states, the complete text is bit-identical; across the three text states, the image path is identical. Automated checks verify all nine cells, direction balance, and exact reuse of no-image controls. Deduplicating shared text-only and blank evaluations yields 3,000 scored probes per model.

For every complete textual state \(s_T\), define the controlled visual effect
\begin{equation}
\Delta_V(s_T)=m(I^+,s_T)-m(I^-,s_T).
\end{equation}
Positive \(\Delta_V(s_T)\) means that changing only the rendered relation moves the Yes margin in the correct direction. The symmetric text effect \(\Delta_T(I)=m(I,s_T^+)-m(I,s_T^-)\) holds the image fixed. Agreement cells \((I^+,s_T^+)\) and \((I^-,s_T^-)\) have unambiguous Yes and No labels; conflict cells have no gold answer and report which source determines the prediction. For a correct agreement decision with \(G^{\rm img}\leq0\), we call \(\Delta_V(s_T)>0\) relation-responsive and separately record the stricter event that the binary answer changes under \(I^+\leftrightarrow I^-\).

\begin{table*}[t]
\centering
{\scriptsize
\setlength{\tabcolsep}{1.3pt}
\begin{tabular*}{\textwidth}{@{\extracolsep{\fill}}lrrrrrrr@{}}
\toprule
Model & Correct \(n\) & C-U \(n\) (rate) & Responsive [95\% CI]
& Answer flip [95\% CI] & Resp. mass & Nonresp. mass & Conflict win \\
\midrule
Qwen & 400 & 305 (76.25) & 100.00 [98.76,100.00]
& 34.43 [29.32,39.92] & 76.25 & 0.00 & 50.00 \\
InternVL & 400 & 331 (82.75) & 81.57 [77.04,85.38]
& 21.75 [17.65,26.51] & 67.50 & 15.25 & 22.25 \\
LLaVA & 310 & 8 (2.58) & 100.00 [67.56,100.00]
& 100.00 [67.56,100.00] & 2.58 & 0.00 & 55.25 \\
Ministral & 400 & 38 (9.50) & 100.00 [90.82,100.00]
& 89.47 [75.87,95.83] & 9.50 & 0.00 & 90.50 \\
Pooled & 1,510 & 682 (45.17) & 91.06 [89.21,92.83]
& 32.11 [29.34,34.81] & 41.13 & 4.04 & 54.50 \\
\bottomrule
\end{tabular*}}
\caption{Factorial decomposition. Correct \(n\) is out of 400 agreement decisions per model. C-U is conditioned on correct decisions; Responsive and Answer flip are conditioned on C-U. Responsive and nonresponsive mass use all correct decisions as denominator. Model-level conditional intervals are Wilson intervals; pooled intervals resample stem ids while retaining all model records.}
\label{tab:supp_factorial_main}
\end{table*}

The pooled correct-uncredited fraction is 45.17 [43.62,46.75], the responsive fraction is 91.06 [89.21,92.83], and the strict answer-flip fraction is 32.11 [29.34,34.81]. The distinction matters: a score can move in the correct direction without crossing the binary decision boundary. The corresponding unweighted model-macro values are 42.77, 95.39, and 61.41. We treat the micro-pooled conditional rates as primary because equal model weighting gives LLaVA's eight and Ministral's 38 C-U decisions the same weight as the much larger Qwen and InternVL subsets. All four models have positive mean \(\Delta_V\) under every textual state. The nondegenerate intervals make the small-denominator uncertainty explicit: LLaVA's two 100\% rates use eight C-U decisions and have lower bounds of 67.56\%, rather than implying a precisely known population rate. The sign result is not confined to numerical ties: every uncredited response for Qwen, LLaVA, and Ministral exceeds 1.0 margin unit; for InternVL, 76.74\% exceed 0.1 and 66.16\% exceed 0.5. These magnitudes are interpreted within model rather than as a cross-backbone scale.

\subsection{Object-Preserving Relation-Neutral Control}

The opposite-relation factorial changes object positions, so we add a matched control that retains the deterministic background, object identities, shapes, sizes, and text. For left/right queries, the two object centers share the same \(x\)-coordinate and remain separated vertically; for above/below, they share the same \(y\)-coordinate and remain separated horizontally. The resulting image contains both objects but neither ordering on the queried axis. For each agreement cell, with correct-answer direction \(d\in\{-1,+1\}\), we measure
\begin{equation}
\Delta_N=d\,[m(I^{\rm rel},s_T)-m(I^{\rm neutral},s_T)].
\end{equation}
Positive \(\Delta_N\) means that the relation-bearing image contributes more support in the agreement direction than general image and object presence. Intervals cluster-bootstrap the 200 stem ids with 5,000 resamples.

\begin{table}[t]
\centering
{\scriptsize
\setlength{\tabcolsep}{2.2pt}
\begin{tabular*}{\columnwidth}{@{\extracolsep{\fill}}lrrrr@{}}
\toprule
Model & Correct \(n\) & C-U \(n\) & \(\Delta_N>0\) [95\% CI] & Mean \(\Delta_N\) [95\% CI] \\
\midrule
Qwen & 400 & 305 & 100.00 [100.00,100.00] & 3.712 [3.569,3.840] \\
InternVL & 400 & 331 & 96.07 [93.87,97.94] & 0.557 [0.518,0.595] \\
LLaVA & 310 & 8 & 75.00 [40.00,100.00] & 0.926 [0.385,1.371] \\
Ministral & 400 & 38 & 100.00 [100.00,100.00] & 4.589 [4.203,4.990] \\
Pooled & 1,510 & 682 & 97.80 [96.71,98.82] & 2.197 [2.091,2.301] \\
\bottomrule
\end{tabular*}}
\caption{Same-object relation-neutral control within correct-but-uncredited agreement decisions. Margins are in model-native units and are not compared across backbones.}
\label{tab:supp_object_neutral}
\end{table}

The pooled rate and its lower confidence bound exceed chance, as do the two models contributing 636 of 682 C-U decisions. Thus, in this controlled domain, a generic response to seeing the named objects does not explain 97.80\% of the pooled contrasts. LLaVA's eight-case interval remains correspondingly wide.

\begin{table}[t]
\centering
{\scriptsize
\setlength{\tabcolsep}{2.0pt}
\begin{tabular*}{\columnwidth}{@{\extracolsep{\fill}}lrrrrrrr@{}}
\toprule
& \multicolumn{3}{c}{Visual effect \(\Delta_V(s_T)\)}
& \multicolumn{3}{c}{Text effect \(\Delta_T(I)\)} & Inter. \\
Model & \(s_T^+\) & \(s_T^-\) & \(s_T^\emptyset\)
& \(I^+\) & \(I^-\) & \(I^\emptyset\) & \\
\midrule
Qwen & 10.684 & 3.538 & 11.688 & 11.141 & 3.995 & 1.138 & 7.146 \\
InternVL & 0.439 & 1.825 & 3.494 & 1.472 & 2.858 & 1.377 & -1.386 \\
LLaVA & 2.791 & 1.986 & 3.052 & 1.781 & 0.975 & 0.290 & 0.806 \\
Ministral & 14.086 & 9.117 & 17.025 & 5.520 & 0.552 & 0.596 & 4.968 \\
\bottomrule
\end{tabular*}}
\caption{Mean conditional source effects in native margin units. Interaction is \(\Delta_V(s_T^+)-\Delta_V(s_T^-)\); magnitudes are interpreted within model.}
\label{tab:supp_factorial_effects}
\end{table}

\begin{table}[t]
\centering
{\scriptsize
\setlength{\tabcolsep}{2.0pt}
\begin{tabular*}{\columnwidth}{@{\extracolsep{\fill}}lrrrr@{}}
\toprule
Model & Dep T/F & D-CC T/F & Corr. VCA & Corr. Conf. \\
\midrule
Qwen & 100.00/50.00 & 100.00/0.00 & 100.00 & 98.83 \\
InternVL & 88.00/15.75 & 87.50/0.00 & 100.00 & 72.10 \\
LLaVA & 74.25/40.75 & 65.75/1.25 & 99.66 & 84.20 \\
Ministral & 99.50/26.00 & 99.50/0.00 & 100.00 & 99.98 \\
\bottomrule
\end{tabular*}}
\caption{Nested source-only validation. T/F denotes image-source TPR and text-source FPR. Correct-decision columns report source AUROC for VCA and confidence.}
\label{tab:supp_factorial_source}
\end{table}

The nested source-only cells reproduce the earlier validation question while the full factorial answers the stronger responsiveness question. Correct- decision VCA source AUROC is 99.66--100.00. Thus VCA accurately identifies marginal source advantage in these controlled inputs, while the factorial demonstrates why marginal advantage and counterfactual response must not be collapsed into one claim.

\subsection{Natural-Image Correspondence}

We next join the same 108 accepted object-swap edits to each model's canonical source query. This is not a clean causal intervention because patch editing can introduce rendering artifacts; its role is to test whether the controlled phenotype also appears in natural benchmark records. Among source-correct records, we stratify by canonical VCA credit and evaluate whether the edited image induces the expected answer flip while both objects remain visible.

\begin{table}[t]
\centering
{\scriptsize
\setlength{\tabcolsep}{2.1pt}
\begin{tabular*}{\columnwidth}{@{\extracolsep{\fill}}lrrrr@{}}
\toprule
Model & Correct \(n\) & C-U \(n\) & C-U rate & Flip+vis. within C-U \\
\midrule
Qwen & 100 & 2 & 2.00 & 50.00 \\
InternVL & 106 & 59 & 55.66 & 22.03 \\
LLaVA & 100 & 6 & 6.00 & 16.67 \\
Ministral & 103 & 33 & 32.04 & 30.30 \\
Pooled & 409 & 100 & 24.45 & 25.00 \\
\bottomrule
\end{tabular*}}
\caption{Natural-image correspondence on three-annotator-accepted edits. C-U means source-correct but uncredited. The pooled flip+visibility interval within C-U is [17.20,33.05].}
\label{tab:supp_natural_uncredited}
\end{table}

The pooled rate shows that a substantial minority of naturally occurring correct-but-uncredited records still respond to a relation edit. The controlled factorial supplies identification; this artifact-sensitive edit set supplies cross-domain correspondence, not proof that every uncredited natural decision contains redundant visual evidence.

\section{Probe Construction and Geometry}

\subsection{Matched Context Construction}

Bounding boxes are used only by the offline renderer for optional stress views; the frozen MLLM never receives boxes, masks, depth, or segmentation labels. The primary image-dependence audit uses no boxes. Table~\ref{tab:supp_context_construction} records the exact matched contexts and whether each enters a headline quantity.

\par\noindent\begin{minipage}{\columnwidth}
\centering
{\scriptsize
\setlength{\tabcolsep}{1.4pt}
\begin{tabular*}{\columnwidth}{@{\extracolsep{\fill}}L{0.22\columnwidth}L{0.54\columnwidth}L{0.20\columnwidth}@{}}
\toprule
Context & Construction & Role \\
\midrule
Original & Unmodified RGB image and fixed spatial query. & Primary support \\
Text-only & Identical query without an image. & Primary control \\
Blank & Same-size fixed mid-gray canvas. & Primary control \\
Mismatch & Fixed query with a same-split image. & Robustness only \\
Relation-erased & Padded subject--object union filled with image mean. & Box stress test \\
Irrelevant-erased & Distant tiles filled outside the protected union. & Box stress test \\
SRC & True and false relations on the same image and pair. & Relation axis \\
Object swap & Object patches moved to exchanged positions. & External check \\
\bottomrule
\end{tabular*}}
\captionof{table}{Exact VCA context construction. Only original, text-only, and blank enter the primary image-dependence gap.}
\label{tab:supp_context_construction}
\end{minipage}\par\medskip

Original and blank supply one image, while text-only supplies none and retains no visual placeholder. Within each backend, the question, native chat template, generation prefix, and answer verbalizers remain fixed within each interface. Qwen uses the explicit system message ``You are a helpful assistant.''; the other three backends add no system message. Qwen, InternVL, and LLaVA use their AutoProcessor templates; Ministral uses MistralCommon. For Qwen, InternVL, and LLaVA, the leading-space verbalizers \texttt{yes/no/true/false} are single tokens with ids 9834/902/830/895; Ministral's native continuation tokens have ids 13059/2649/5876/11339. No free-form or chain-of-thought decoding is invoked. Blank preserves source dimensions before native transforms. Prompt/BOS tokens are masked, and loss averages only continuation tokens.

\subsection{GSR-COCO Pairing}

GSR-COCO is converted from the two-object COCO-Spatial-Two split using the supplied caption-option metadata. Each option is parsed into subject, relation, and object fields and converted to a literal yes/no statement. The metadata order fixes the true option; its paired alternative supplies the false statement. We preserve image ids, option ids, relations, and split names and do not rebalance relations or filter samples using model outputs. This produces 880 statements in 440 fixed same-image pairs before any MLLM is scored. The paired construction gives exactly 440 true and 440 false statements; relation counts are 279 each for left/right and 161 each for above/below.

\subsection{Erasure Geometry}

RelE removes subject--object support rather than isolating the relation; Table~\ref{tab:supp_erasure_geometry} quantifies its scope. The padded union covers both boxes and most of the image. IrrE fills outside the protected union; zero-overlap is harder on VSR because large boxes leave little background.

\begin{table}[t]
\centering
{\scriptsize
\setlength{\tabcolsep}{2.0pt}
\begin{tabular*}{\columnwidth}{@{\extracolsep{\fill}}lrrrrr@{}}
\toprule
Data & $N$ & Rel mean & Rel med. & Irr mean & Irr zero \\
\midrule
VSR & 1196 & 74.8 & 77.6 & 7.1 & 59.9 \\
GSR-COCO & 880 & 60.8 & 59.7 & 11.0 & 83.4 \\
\bottomrule
\end{tabular*}}
\caption{Erasure geometry as percentage of image area. Zero overlap is the percentage of IrrE views with no erased pixels inside the protected union.}
\label{tab:supp_erasure_geometry}
\end{table}

\section{Decision Survival and Outcome Decomposition}

\subsection{Decision Survival Under No-Image Controls}

To separate incremental support from answer-level necessity, we condition on \(D_i=1\) and record whether each no-image control preserves the original binary prediction. ``Both same'' means that neither control changes the answer; ``Any changes'' is its complement.

\par\noindent\begin{minipage}{\columnwidth}
\centering
{\scriptsize
\setlength{\tabcolsep}{1.7pt}
\begin{tabular*}{\columnwidth}{@{\extracolsep{\fill}}llrrrr@{}}
\toprule
Model & Data & Text same & Blank same & Both same & Any changes \\
\midrule
Qwen & VSR & 26.80 & 25.45 & 20.72 & 79.28 \\
Qwen & GSR & 42.50 & 44.31 & 39.44 & 60.56 \\
InternVL & VSR & 86.94 & 80.34 & 78.79 & 21.21 \\
InternVL & GSR & 75.33 & 75.60 & 75.33 & 24.67 \\
LLaVA & VSR & 70.34 & 1.57 & 0.10 & 99.90 \\
LLaVA & GSR & 72.71 & 19.89 & 1.76 & 98.24 \\
Ministral & VSR & 62.20 & 70.67 & 54.87 & 45.13 \\
Ministral & GSR & 55.32 & 55.51 & 51.90 & 48.10 \\
\bottomrule
\end{tabular*}}
\captionof{table}{Decision survival among credited samples (\%).}
\label{tab:supp_decision_survival}
\end{minipage}\par\medskip

The wide range reflects model- and context-specific answer priors. It also makes the construct boundary explicit: positive VCA credit means that removing image content lowers support, not necessarily that the image is required to preserve the binary answer. Decision survival is therefore reported as a phenotype rather than folded into the credit rule.

\subsection{Answer-Direction Stress Tests}

Strict isolation additionally requires both no-image controls to favor the opposite original decision. Table~\ref{tab:supp_prediction_direction} shows why this is unsuitable as the primary definition. Its answer-direction behavior is extreme and model-dependent: it nearly vanishes on Qwen and LLaVA negative predictions, but is high on InternVL negative predictions because InternVL has a different response prior. The non-negating Dep test still reveals substantial directional differences, but it no longer equates missing evidence with the opposite proposition.

\par\noindent\begin{minipage}{\columnwidth}
\centering
{\scriptsize
\renewcommand{\arraystretch}{0.88}
\setlength{\tabcolsep}{1.5pt}
\begin{tabular*}{\columnwidth}{@{\extracolsep{\fill}}lllrrrrr@{}}
\toprule
Model & Data & Prediction & \(N\) & Acc & Dep & D-CC & Strict \\
\midrule
Qwen & VSR & Yes & 672 & 80.51 & 98.36 & 79.32 & 90.48 \\
Qwen & VSR & No & 577 & 78.16 & 39.34 & 34.32 & 0.00 \\
Qwen & GSR-COCO & Yes & 402 & 94.78 & 99.00 & 93.78 & 93.53 \\
Qwen & GSR-COCO & No & 478 & 87.66 & 67.36 & 65.06 & 0.21 \\
\midrule
InternVL & VSR & Yes & 1157 & 56.53 & 54.19 & 36.47 & 0.69 \\
InternVL & VSR & No & 92 & 85.87 & 92.39 & 78.26 & 80.43 \\
InternVL & GSR-COCO & Yes & 784 & 54.72 & 36.22 & 28.06 & 0.00 \\
InternVL & GSR-COCO & No & 96 & 88.54 & 96.88 & 86.46 & 95.83 \\
\midrule
LLaVA & VSR & Yes & 997 & 64.59 & 94.18 & 63.19 & 26.98 \\
LLaVA & VSR & No & 252 & 90.87 & 5.95 & 5.56 & 0.00 \\
LLaVA & GSR-COCO & Yes & 519 & 77.07 & 88.25 & 73.22 & 9.83 \\
LLaVA & GSR-COCO & No & 361 & 88.92 & 30.47 & 30.47 & 0.28 \\
\midrule
Ministral & VSR & Yes & 1037 & 60.75 & 59.50 & 44.36 & 5.50 \\
Ministral & VSR & No & 212 & 82.55 & 82.08 & 68.87 & 55.19 \\
Ministral & GSR-COCO & Yes & 641 & 64.74 & 45.87 & 43.37 & 0.00 \\
Ministral & GSR-COCO & No & 239 & 89.54 & 97.07 & 87.45 & 89.96 \\
\bottomrule
\end{tabular*}}
\captionof{table}{Results by original prediction. Strict requires opposite-answer isolation under both no-image controls.}
\label{tab:supp_prediction_direction}
\end{minipage}\par\smallskip

\par\noindent\begin{minipage}{\columnwidth}
\centering
{\scriptsize
\renewcommand{\arraystretch}{0.96}
\setlength{\tabcolsep}{1.5pt}
\begin{tabular*}{\columnwidth}{@{\extracolsep{\fill}}lllrrrrr@{}}
\toprule
Model & Data & Gold answer & \(N\) & Acc & Dep & D-CC & Strict \\
\midrule
Qwen & VSR & Yes & 667 & 81.11 & 84.26 & 79.91 & 73.01 \\
Qwen & VSR & No & 582 & 77.49 & 56.01 & 34.02 & 20.79 \\
Qwen & GSR-COCO & Yes & 440 & 86.59 & 88.18 & 85.68 & 80.68 \\
Qwen & GSR-COCO & No & 440 & 95.23 & 75.45 & 70.68 & 5.00 \\
\midrule
InternVL & VSR & Yes & 667 & 98.05 & 65.22 & 63.27 & 3.00 \\
InternVL & VSR & No & 582 & 13.57 & 47.59 & 12.37 & 10.65 \\
InternVL & GSR-COCO & Yes & 440 & 97.50 & 52.27 & 50.00 & 2.27 \\
InternVL & GSR-COCO & No & 440 & 19.32 & 33.41 & 18.86 & 18.64 \\
\midrule
LLaVA & VSR & Yes & 667 & 96.55 & 94.60 & 94.45 & 28.04 \\
LLaVA & VSR & No & 582 & 39.35 & 55.50 & 2.41 & 14.09 \\
LLaVA & GSR-COCO & Yes & 440 & 90.91 & 86.36 & 86.36 & 8.18 \\
LLaVA & GSR-COCO & No & 440 & 72.95 & 42.73 & 25.00 & 3.64 \\
\midrule
Ministral & VSR & Yes & 667 & 94.45 & 73.16 & 68.97 & 9.30 \\
Ministral & VSR & No & 582 & 30.07 & 52.06 & 25.09 & 19.24 \\
Ministral & GSR-COCO & Yes & 440 & 94.32 & 68.41 & 63.18 & 5.00 \\
Ministral & GSR-COCO & No & 440 & 48.64 & 51.14 & 47.50 & 43.86 \\
\bottomrule
\end{tabular*}}
\captionof{table}{Results stratified by the benchmark answer.}
\label{tab:supp_gold_direction}
\end{minipage}\par\smallskip

\par\noindent\begin{minipage}{\columnwidth}
\centering
{\scriptsize
\renewcommand{\arraystretch}{0.92}
\setlength{\tabcolsep}{1.2pt}
\begin{tabular*}{\columnwidth}{@{\extracolsep{\fill}}llrrrr@{}}
\toprule
& & \multicolumn{2}{c}{Prediction-balanced} &
\multicolumn{2}{c}{Gold-balanced} \\
Model & Data & D-CC & C-U & D-CC & C-U \\
\midrule
Qwen & VSR & 56.82 & 22.52 & 56.97 & 22.33 \\
Qwen & GSR-COCO & 79.42 & 11.80 & 78.18 & 12.73 \\
InternVL & VSR & 57.37 & 13.84 & 37.82 & 17.99 \\
InternVL & GSR-COCO & 57.26 & 14.37 & 34.43 & 23.98 \\
LLaVA & VSR & 34.38 & 43.36 & 48.43 & 19.52 \\
LLaVA & GSR-COCO & 51.84 & 31.15 & 55.68 & 26.25 \\
Ministral & VSR & 56.62 & 15.04 & 47.03 & 15.23 \\
Ministral & GSR-COCO & 65.41 & 11.73 & 55.34 & 16.14 \\
\bottomrule
\end{tabular*}}
\captionof{table}{Equal-weight direction sensitivity derived from the preceding strata. Prediction-balanced averages original Yes and No predictions; gold-balanced averages benchmark Yes and No answers. C-U is balanced Acc\(-\)D-CC. These stress estimands do not replace benchmark-distribution rates.}
\label{tab:supp_direction_balanced}
\end{minipage}\par\smallskip

Gold-direction balancing preserves the headline 12.73--26.25\% C-U range and the VSR D-CC ordering. Prediction-direction balancing yields 11.73--43.36\% C-U and changes the VSR ordering, demonstrating that benchmark-distribution rates are not direction-invariant model abilities.

\subsection{Audit Outcome Decomposition}

A positive image gap and correctness are logically independent. Table~\ref{tab:supp_audit_quadrants} partitions every run into the four resulting events. Correct/not-dependent is the portion of ordinary accuracy withheld by VCA; wrong/dependent is the portion where the image influenced the model but did not produce the benchmark answer. Credited precision reports correctness conditional on $D_i=1$.

\begin{table*}[t]
\centering
{\small
\setlength{\tabcolsep}{3.2pt}
\begin{tabular*}{\textwidth}{@{\extracolsep{\fill}}llrrrrr@{}}
\toprule
Model & Data & Correct+D & Correct+$\neg$D & Wrong+D & Wrong+$\neg$D & Credited precision \\
\midrule
Qwen & VSR & 58.53 & 20.90 & 12.57 & 8.01 & 82.32 \\
Qwen & GSR-COCO & 78.18 & 12.73 & 3.64 & 5.45 & 95.56 \\
InternVL & VSR & 39.55 & 19.14 & 17.45 & 23.86 & 69.38 \\
InternVL & GSR-COCO & 34.43 & 23.98 & 8.41 & 33.18 & 80.37 \\
LLaVA & VSR & 51.56 & 18.33 & 24.82 & 5.28 & 67.51 \\
LLaVA & GSR-COCO & 55.68 & 26.25 & 8.86 & 9.20 & 86.27 \\
Ministral & VSR & 48.52 & 15.93 & 14.81 & 20.74 & 76.61 \\
Ministral & GSR-COCO & 55.34 & 16.14 & 4.43 & 24.09 & 92.59 \\
\bottomrule
\end{tabular*}}
\caption{Complete audit taxonomy. The four middle columns partition all samples and therefore sum to 100 within each row. Values are percentages.}
\label{tab:supp_audit_quadrants}
\end{table*}

\subsection{Continuous Raw-Gap Distributions}

Binary Dep records the gap sign; its magnitude distinguishes near-zero boundary cases from strong per-item evidence. Table~\ref{tab:supp_gap_quantiles} shows broad overlap across models and datasets rather than a degenerate mass at zero. In particular, InternVL/GSR-COCO has a negative median and mean but a positive upper quartile, explaining how its Dep rate can remain 42.84\%.

\begin{table*}[t]
\centering
{\small
\setlength{\tabcolsep}{3.5pt}
\begin{tabular*}{\textwidth}{@{\extracolsep{\fill}}llrrrrrr@{}}
\toprule
Model & Data & Mean & P10 & P25 & Median & P75 & P90 \\
\midrule
Qwen & VSR & 2.476 & $-1.781$ & $-0.312$ & 2.062 & 4.969 & 7.475 \\
Qwen & GSR-COCO & 2.953 & $-0.688$ & 0.781 & 2.797 & 4.883 & 7.284 \\
InternVL & VSR & 0.297 & $-1.826$ & $-0.740$ & 0.233 & 1.318 & 2.272 \\
InternVL & GSR-COCO & $-0.264$ & $-2.729$ & $-1.629$ & $-0.379$ & 1.028 & 2.316 \\
LLaVA & VSR & 1.072 & $-1.609$ & 0.103 & 1.254 & 2.277 & 3.219 \\
LLaVA & GSR-COCO & 0.651 & $-1.362$ & $-0.438$ & 0.499 & 1.894 & 2.872 \\
Ministral & VSR & 0.898 & $-1.875$ & $-0.672$ & 0.688 & 2.312 & 3.781 \\
Ministral & GSR-COCO & 0.605 & $-2.361$ & $-1.281$ & 0.656 & 2.375 & 3.375 \\
\bottomrule
\end{tabular*}}
\caption{Distribution of the unmodified raw image gap. Quantiles are computed across samples before thresholding.}
\label{tab:supp_gap_quantiles}
\end{table*}

The sign threshold is not supported by a large numerical boundary mass. Across the eight model--dataset runs, only 0.08--0.72\% of items satisfy \(|G^{\mathrm{img}}|\leq 0.01\), and 2.40--5.91\% satisfy \(|G^{\mathrm{img}}|\leq 0.1\). This complements the positive-threshold sweep: the headline split is not determined by many near-zero ties.

\section{Relation Consistency, Joint Audit, and Interaction}

\par\noindent\begin{minipage}{\columnwidth}
\centering
{\scriptsize
\setlength{\tabcolsep}{1.0pt}
\begin{tabular*}{\columnwidth}{@{\extracolsep{\fill}}lrrrrrrrr@{}}
\toprule
Model & Groups & Rel.$_o$ & Rel.$_t$ & Rel.$_b$ & PairCred. & Pair D & Joint & Mean int. \\
\midrule
Qwen & 440 & 82.50 & 2.95 & 1.14 & 92.95 & 70.23 & 63.64 & 6.260 \\
InternVL & 440 & 19.09 & 0.00 & 2.27 & 62.05 & 20.23 & 10.68 & 1.193 \\
LLaVA & 440 & 66.14 & 5.68 & 0.68 & 88.86 & 40.45 & 23.86 & 2.476 \\
Ministral & 440 & 44.32 & 2.73 & 0.45 & 85.45 & 41.14 & 34.32 & 2.946 \\
\bottomrule
\end{tabular*}}
\captionof{table}{GSR-COCO relation decomposition. Joint is Rel.$_o\wedge$Pair D; Mean int. is \(G^{\rm int}\).}
\label{tab:supp_relation_axis}
\end{minipage}\par\smallskip

Mean \(H_o/H_t/H_b\) is \(6.529/0.084/0.014\) for Qwen, \(1.951/0.135/0.033\) for InternVL, \(2.656/0.039/0.054\) for LLaVA, and \(3.372/0.021/{-}0.016\) for Ministral. The 5,000-pair bootstrap intervals for mean \(G^{\rm int}\) are [5.867,6.643], [0.943,1.435], [2.297,2.665], and [2.687,3.201], respectively. Thus each backbone shows positive average image--relation interaction, while per-pair PairCredit retains heterogeneity. Pair D alone does not require correct relation discrimination; Joint combines Pair D with Rel.$_o$. The resulting implication for Joint is definitional; PairCredit and mean \(G^{\rm int}\) over all groups are the empirical quantities that retain heterogeneity outside the Joint subset.

\begin{table*}[t]
{\centering\large\bfseries Relation-Wise VSR Results\par}
\vspace{2pt}
\noindent Table~\ref{tab:supp_vsr_relations} reports all nine VSR relation families, exposing substantial relation-wise variation in dependence and credited precision.\par
\vspace{2pt}
\centering
{\scriptsize
\renewcommand{\arraystretch}{0.88}
\setlength{\tabcolsep}{2.2pt}
\begin{tabular*}{0.998\textwidth}{@{\extracolsep{\fill}}llrrrrrr@{}}
\toprule
Model & Relation & $N$ & Acc & Dep & D-CC & Cred. prec. & Mean gap \\
\midrule
Qwen & above & 195 & 77.95 & 72.82 & 61.03 & 83.80 & 2.609 \\
Qwen & behind & 150 & 72.00 & 60.67 & 46.67 & 76.92 & 1.669 \\
Qwen & below & 231 & 78.79 & 64.94 & 51.95 & 80.00 & 2.861 \\
Qwen & contains & 66 & 75.76 & 54.55 & 40.91 & 75.00 & 1.960 \\
Qwen & in\_front\_of & 142 & 80.28 & 71.83 & 59.15 & 82.35 & 2.051 \\
Qwen & inside & 103 & 85.44 & 71.84 & 66.02 & 91.89 & 1.946 \\
Qwen & left\_of & 136 & 80.15 & 86.76 & 69.12 & 79.66 & 3.088 \\
Qwen & on & 117 & 85.47 & 70.09 & 60.68 & 86.59 & 2.075 \\
Qwen & right\_of & 109 & 81.65 & 85.32 & 71.56 & 83.87 & 3.559 \\
\midrule
InternVL & above & 195 & 61.03 & 40.00 & 28.72 & 71.79 & $-0.318$ \\
InternVL & behind & 150 & 58.00 & 58.00 & 40.67 & 70.11 & 0.349 \\
InternVL & below & 231 & 54.55 & 55.41 & 35.06 & 63.28 & 0.122 \\
InternVL & contains & 66 & 53.03 & 50.00 & 39.39 & 78.79 & 0.212 \\
InternVL & in\_front\_of & 142 & 58.45 & 66.20 & 38.03 & 57.45 & 0.428 \\
InternVL & inside & 103 & 70.87 & 80.58 & 61.17 & 75.90 & 1.667 \\
InternVL & left\_of & 136 & 57.35 & 36.03 & 31.62 & 87.76 & $-0.405$ \\
InternVL & on & 117 & 62.39 & 70.09 & 52.14 & 74.39 & 0.849 \\
InternVL & right\_of & 109 & 54.13 & 71.56 & 44.95 & 62.82 & 0.568 \\
\midrule
LLaVA & above & 195 & 68.72 & 85.64 & 58.46 & 68.26 & 1.408 \\
LLaVA & behind & 150 & 67.33 & 82.00 & 54.00 & 65.85 & 1.004 \\
LLaVA & below & 231 & 62.34 & 83.12 & 51.52 & 61.98 & 1.065 \\
LLaVA & contains & 66 & 68.18 & 68.18 & 39.39 & 57.78 & 1.220 \\
LLaVA & in\_front\_of & 142 & 66.90 & 83.80 & 55.63 & 66.39 & 0.781 \\
LLaVA & inside & 103 & 87.38 & 62.14 & 49.51 & 79.69 & 0.851 \\
LLaVA & left\_of & 136 & 68.38 & 70.59 & 50.00 & 70.83 & 1.144 \\
LLaVA & on & 117 & 70.09 & 62.39 & 38.46 & 61.64 & 0.723 \\
LLaVA & right\_of & 109 & 81.65 & 68.81 & 55.96 & 81.33 & 1.361 \\
\midrule
Ministral & above & 195 & 71.28 & 75.38 & 58.46 & 77.55 & 1.109 \\
Ministral & behind & 150 & 58.67 & 51.33 & 36.67 & 71.43 & 0.068 \\
Ministral & below & 231 & 58.01 & 47.62 & 35.06 & 73.64 & $-0.072$ \\
Ministral & contains & 66 & 71.21 & 74.24 & 59.09 & 79.59 & 2.048 \\
Ministral & in\_front\_of & 142 & 64.79 & 79.58 & 54.23 & 68.14 & 1.390 \\
Ministral & inside & 103 & 81.55 & 76.70 & 66.99 & 87.34 & 2.157 \\
Ministral & left\_of & 136 & 56.62 & 46.32 & 39.71 & 85.71 & 0.075 \\
Ministral & on & 117 & 67.52 & 82.91 & 59.83 & 72.16 & 2.753 \\
Ministral & right\_of & 109 & 59.63 & 51.38 & 43.12 & 83.93 & 0.224 \\
\bottomrule
\end{tabular*}}
\caption{Raw VCA results for every VSR relation family. Values except $N$ and Mean gap are percentages.}
\label{tab:supp_vsr_relations}
\end{table*}

\section{Human-Audited Edit Construction}

Fixed geometry selected 114 of 440 source pairs without using model outputs. Three non-author volunteers independently assessed shuffled, blinded batches with repeat checks, and majority consensus retained 108 edits. Every model is evaluated on this same accepted set.

\section{Same-Record and Independent Edit Correspondence}

We test whether the headline dual-control VCA gap ranks two prespecified relation behaviors. SRC is a same-record consistency check rather than an independent target: for each pair, the predictor is the raw gap of the canonical true-relation statement; the false statement enters only the same-image consistency target, and statement gaps are not aggregated. Each edit uses six binary probes: source-image \(q_r/q_{r^{-1}}\) (targets yes/no), edited-image \(q_r/q_{r^{-1}}\) (no/yes), and edited-image subject/object presence (yes/yes). Let \(z(p)\) indicate a yes prediction. Because every accepted edit has human-confirmed object visibility, the relation-only target is \((1-z(q_r,I'))z(q_{r^{-1}},I')\). The primary model-checked target is \((1-z(q_r,I'))z(q_{r^{-1}},I')z(q_s^{\rm pres},I')z(q_o^{\rm pres},I')\); the strict sensitivity additionally requires \(z(q_r,I)(1-z(q_{r^{-1}},I))\). Thus the primary target contains no source condition. Neither target defines or tunes \(G^{\mathrm{img}}\). All four models evaluate each of the 108 accepted edits (432 correlated records).

\par\noindent\begin{minipage}{\columnwidth}
\centering
{\scriptsize
\setlength{\tabcolsep}{1.8pt}
\begin{tabular*}{\columnwidth}{@{\extracolsep{\fill}}lllrrrr@{}}
\toprule
Check & Model & \(N\) & Base rate & AUROC & AUPRC & H--L \\
\midrule
Swap strict & Qwen & 108 & 62.96 & 83.99 & 85.86 & +63.89 \\
Swap strict & InternVL & 108 & 8.33 & 50.51 & 10.05 & +2.78 \\
Swap strict & LLaVA & 108 & 51.85 & 90.56 & 91.80 & +83.33 \\
Swap strict & Ministral & 108 & 26.85 & 76.52 & 57.28 & +41.67 \\
Swap strict & Rank-pooled & 432 & 37.50 & 74.85 & 59.01 & -- \\
Swap flip+vis. & Rank-pooled & 432 & 48.15 & 67.93 & 63.39 & -- \\
\midrule
SRC & Qwen & 440 & 82.50 & 92.29 & 97.73 & +44.53 \\
SRC & InternVL & 440 & 19.09 & 55.55 & 21.53 & +8.22 \\
SRC & LLaVA & 440 & 66.14 & 78.71 & 87.45 & +49.38 \\
SRC & Ministral & 440 & 44.32 & 65.34 & 61.64 & +29.45 \\
SRC & Macro (4 models) & -- & 53.01 & 72.97 & 67.09 & +32.88 \\
\bottomrule
\end{tabular*}}
\captionof{table}{Standalone predictive validity of the headline dual-control VCA gap. Strict requires original-pair correctness plus edited flip+visibility; flip+vis. omits the original-pair condition. Rank pooling is performed within model. H--L is the high-minus-low tertile success difference.}
\label{tab:supp_external_validity}
\end{minipage}\par\medskip

SRC macro rows average metrics computed separately within each backbone; raw scores are never ranked across models. The swap pooled rows first convert each predictor to within-model percentile ranks. Its model-specific AUROCs expose substantial heterogeneity, especially the low strict-flip base rate for InternVL. Swap high--low intervals use edit-id cluster bootstrap, and SRC intervals use relation-pair clusters. The 67.93 row targets standalone flip+visibility, 74.85 targets the stricter composite, and the cross-fitted analysis below tests incremental prediction after covariate control.

\paragraph{Confidence--VCA discordance.} We split confidence at the median separately within each model and dataset, then cross it with the untuned event \(G^{\mathrm{img}}>0\). This descriptive split covers all 8,516 test records and joins the same source records to all 432 model--edit outcomes.

\par\noindent\begin{minipage}{\columnwidth}
\centering
{\scriptsize
\setlength{\tabcolsep}{1.2pt}
\begin{tabular*}{\columnwidth}{@{\extracolsep{\fill}}lrrrrrr@{}}
\toprule
Source quadrant & Test \(N\) & Share & Acc. & Swap \(N\) & Flip+vis. & Strict \\
\midrule
High conf., credited & 3,715 & 43.62 & 86.11 & 293 & 56.31 & 49.15 \\
High conf., uncredited & 573 & 6.73 & 67.36 & 43 & 20.93 & 6.98 \\
Low conf., credited & 1,821 & 21.38 & 68.31 & 46 & 43.48 & 26.09 \\
Low conf., uncredited & 2,407 & 28.26 & 51.43 & 50 & 28.00 & 6.00 \\
\bottomrule
\end{tabular*}}
\captionof{table}{Systematic confidence--VCA discordance. Test columns pool the two complete benchmarks; swap columns report human-audited edited behavior. Values other than counts are percentages.}
\label{tab:supp_confidence_discordance}
\end{minipage}\par\medskip

High-confidence uncredited cases occur in every model--dataset combination (0.23--18.52\% of its records) and span all evaluated relation families. Their edited outcomes differ sharply from high-confidence credited cases, showing that original-context certainty and matched-control image support are behaviorally distinct signals.

\paragraph{Direct shared-margin control.} Because strict original-image relation consistency (Rel.$_o$) and its predictor share the canonical true statement, we directly compare the gap with that statement's original margin. The original margin reaches 75.83 macro AUROC for Rel.$_o$, versus 72.97 for the gap, and their cross-validated combination changes AUROC by $-0.31$. Rel.$_o$ therefore serves as a same-record convergent check. For a false-only target conditioned on the true statement being accepted, the gap retains direct ranking and a positive increment beyond the conditional baseline.

\par\noindent\begin{minipage}{\columnwidth}
\centering
{\scriptsize
\setlength{\tabcolsep}{1.1pt}
\begin{tabular*}{\columnwidth}{@{\extracolsep{\fill}}lrrrr@{}}
\toprule
Target & True m. & Gap & Cond. & CV $\Delta$ \\
\midrule
Rel. consistency & 75.83 & 72.97 & -- & $-0.31$ [$-0.62$,$-0.12$] \\
False-only & -- & 66.41 & 44.88 & $+0.99$ [$+0.49$,$+1.36$] \\
\bottomrule
\end{tabular*}}
\captionof{table}{Macro AUROC controls for predictor--target score sharing. Conditional gap compares within true-margin bins; CV increment adds the raw gap to the true margin in repeated out-of-fold logistic regression.}
\label{tab:supp_src_margin_control}
\end{minipage}\par\medskip

\paragraph{Independent edit outcomes by audit stratum.} We join each source-image Joint decision to an independently judged edited-image outcome requiring relation-flip recognition and visibility of both objects. The target contains no source-pair consistency term and therefore evaluates transfer beyond the source audit.

\par\noindent\begin{minipage}{\columnwidth}
\centering
{\scriptsize
\setlength{\tabcolsep}{0.5pt}
\begin{tabular*}{\columnwidth}{@{\extracolsep{\fill}}lrrrrrr@{}}
\toprule
Model & R/J \(n\) & Rel. & +Vis. & Rel.$_o$ & Joint & $\Delta_{\rm PairD}$ \\
\midrule
Qwen & 95/79 & 84.26 & 67.59 & 71.58 & 78.48 & $+6.90$ \\
InternVL & 25/15 & 25.93 & 25.93 & 36.00 & 33.33 & $-2.67$ \\
LLaVA & 80/30 & 62.96 & 57.41 & 70.00 & 83.33 & $+13.33$ \\
Ministral & 58/46 & 45.37 & 41.67 & 50.00 & 56.52 & $+6.52$ \\
Macro & 258/170 & 54.63 & 48.15 & 56.89 & 62.92 & $+6.02$ \\
\bottomrule
\end{tabular*}}
\captionof{table}{Component audit on 108 accepted edits per model. R/J \(n\) gives Rel.$_o$/Joint counts. Rel. tests the edited true/false pair after human visibility audit; +Vis. additionally requires both model object-presence responses. Rel.$_o$/Joint use +Vis. and exclude source consistency.}
\label{tab:supp_joint_failure_audit}
\end{minipage}\par\medskip

Relation-only success is 54.63\% overall and 69.47\% within Joint ($+14.84$ [$+8.24$,$+22.24$]); adding the model's two object-presence responses gives 48.15\% and 62.92\%, respectively. Thus 6.48 points of the overall composite failure arise from model visibility responses despite human confirmation that both objects remain visible. Joint raises the full composite by $+14.77$ [$+7.91$,$+22.30$]. Within Rel.$_o$, the Pair D stratum is $+6.02$ [$+0.10$,$+12.16$] points higher; Qwen, LLaVA, and Ministral are positive, whereas InternVL is $-2.67$ [$-20.33$,$+13.16$]. A leave-one-edit-out model restricted to Rel.$_o$ and controlling relation, model, original direction, and within-model confidence rank changes AUROC from 82.01 to 82.03 when Pair D is added: $+0.03$ [$-1.41$,$+1.41$]. The results establish total Joint enrichment and descriptive Pair D stratification; the adjusted Pair D increment remains unresolved. The independent target fails in 37.08\% of Joint records, defining the observed transfer boundary.

\paragraph{Edit-grouped incremental control.} For every held-out edit, we exclude all four associated model records, fit on the other edits, and predict the held-out records. The baseline contains model and relation indicators, original answer direction, and within-model confidence rank. The augmented model adds the within-model rank of the headline \(G^{\mathrm{img}}=r_o-\max(r_t,r_b)\); no erasure or mismatch score enters. Bootstrap intervals resample edit ids and retain their model records.

\par\noindent\begin{minipage}{\columnwidth}
\centering
{\scriptsize
\setlength{\tabcolsep}{1.0pt}
\begin{tabular*}{\columnwidth}{@{\extracolsep{\fill}}lrrrr@{}}
\toprule
Target & Edits/records & Baseline & +VCA & $\Delta$ [95\% CI] \\
\midrule
Flip+visibility & 108/432 & 78.21 & 79.32 & $+1.10$ [$-0.05$,$+2.26$] \\
Strict composite & 108/432 & 89.77 & 90.14 & $+0.37$ [$-0.21$,$+0.96$] \\
Flip given orig. pair & 100/259 & 82.13 & 82.25 & $+0.12$ [$-0.96$,$+1.17$] \\
\bottomrule
\end{tabular*}}
\captionof{table}{Leave-one-edit-out AUROC. The first row isolates edited relation recognition and visibility; the strict row also includes original-pair correctness.}
\label{tab:supp_swap_incremental}
\end{minipage}\par\medskip

The default flip+visibility increment is $+1.10$ [$-0.05$,$+2.26$], and all four model-specific increments are positive at $+1.26$ to $+4.06$. The strict composite is strongly predicted by original support. InternVL illustrates the distinction: 105/108 edits retain both objects, 28/108 produce edited-pair consistency, but only 25/108 have an originally consistent pair and 9/108 pass the strict composite.

Ridge values from $0.1$ to $10$ give gains of $+0.93$ to $+1.18$, with all intervals crossing zero. Replacing confidence plus direction by the within-model signed-margin rank gives $+1.55$ [$+0.37$,$+2.67$]. Leave-one-model-out gains range from $+0.78$ to $+1.30$, with all intervals crossing zero. Horizontal and vertical increments are $+2.03$ [$-0.19$,$+4.31$] and $-1.16$ [$-2.47$,$+0.07$], respectively. The audit therefore shows a positive cross-model tendency together with axis heterogeneity.

The target-supervised regressions above diagnose external validity and do not alter the label-free VCA definition. Confidence needs only the original score; headline VCA uses exactly three scored contexts per item (original, text-only, and blank), versus one for ordinary accuracy. The full two-verbalizer sensitivity check uses six. SRC reuses these GSR-COCO statement scores. Backend-dependent batching precludes a portable wall-clock multiplier, so we report context counts.

\section{Held-Out Benchmark Filtering}

We adapt the closest continuous per-question audit, Korkut et al.'s (2026) Shortcut Score $S(q)=T(q)-V(q)$, to binary GSR-COCO. For each reference model, $T$ is correct-and-confident text-only evidence and $V$ is the gold-answer probability gain from text-only to the original image, normalized by $1-1/2$. Because this comparison has no separate text-only LLM panel, $T$ uses the reference VLMs' text-only branches; this is a close binary-answer adaptation rather than an exact reproduction of their LLM panel. We also adapt VisNec's core score as the gold-answer loss reduction from the text-only control to the original image. Exact VisNec masks visual-token attention in a model-specific blind pass and adds semantic clustering; our cached-control version is therefore denoted VisNec-style rather than an exact reproduction.

For a label-free benchmark-level use of VCA, dependence gaps are rank-normalized within each reference model before averaging. Every filter is constructed from three architectures and evaluated on the held-out fourth architecture. We average scores within the 440 supplied true/false relation groups, retain both questions in each selected group, and compare equal coverage rather than tune a threshold. Exact score ties are broken deterministically by relation-group id before percentile ranking. The confidence panel applies the same group averaging and model-wise rank normalization to original-answer confidence. Blind Gap is text-only accuracy minus 50\% chance; Visual Gain is original-image accuracy minus text-only accuracy. Because retaining a relation group preserves one true and one false statement, text-only accuracy is structurally constrained near 50\% when a model gives both members the same answer; Blind Gap is reported for comparability, while accuracy, Visual Gain, and Dep. carry the filtering interpretation.

\par\noindent\begin{minipage}{\columnwidth}
\centering
{\scriptsize
\setlength{\tabcolsep}{2.0pt}
\begin{tabular*}{\columnwidth}{@{\extracolsep{\fill}}clrrrr@{}}
\toprule
Keep & Filter & Acc. & Blind gap & Visual gain & Dep. \\
\midrule
100\% & Full benchmark & 75.68 & $-0.40$ & 26.08 & 62.24 \\
25\% & Shortcut Score & 85.11 & 0.00 & 35.11 & 73.64 \\
25\% & VisNec-style & 82.05 & 0.00 & 32.05 & 67.73 \\
25\% & Confidence panel & 84.89 & 0.11 & 34.77 & 71.93 \\
25\% & VCA panel & \textbf{85.23} & 0.57 & 34.66 & \textbf{74.43} \\
50\% & Shortcut Score & \textbf{83.98} & 0.06 & \textbf{33.92} & \textbf{71.70} \\
50\% & VisNec-style & 79.77 & $-0.74$ & 30.51 & 66.65 \\
50\% & Confidence panel & 82.61 & $-0.28$ & 32.90 & 70.91 \\
50\% & VCA panel & 83.52 & $-0.11$ & 33.64 & 70.74 \\
75\% & Shortcut Score & \textbf{80.68} & $-0.38$ & \textbf{31.06} & \textbf{67.12} \\
75\% & VisNec-style & 78.03 & $-0.53$ & 28.56 & 64.43 \\
75\% & Confidence panel & 79.81 & $-0.34$ & 30.15 & 66.63 \\
75\% & VCA panel & 79.55 & $-0.19$ & 29.73 & 65.76 \\
\bottomrule
\end{tabular*}}
\captionof{table}{Macro leave-one-model-out filtering on GSR-COCO. Shortcut Score and VisNec-style use gold labels; confidence and the VCA dependence panel do not. Dep. is the held-out model's prediction-aligned dependence rate and is unused by all filters.}
\label{tab:supp_heldout_filtering}
\end{minipage}\par\medskip

At the primary 50\% coverage, 5,000 relation-group bootstrap resamples give VCA-minus-Shortcut differences of $-0.45$ [$-1.64$,$+0.64$] for accuracy, $-0.17$ [$-0.73$,$+0.43$] for Blind Gap, and $-0.28$ [$-1.70$,$+1.01$] for Visual Gain. Against confidence, the corresponding differences are $+0.91$ [$-0.15$,$+1.98$], $+0.17$ [$-0.43$,$+0.77$], and $+0.74$ [$-0.45$,$+1.95$]. Against VisNec-style, they are $+3.75$ [$+2.19$,$+5.37$], $+0.62$ [$-0.23$,$+1.52$], and $+3.12$ [$+1.32$,$+4.91$]. Thus the VCA dependence panel improves accuracy and Visual Gain over the VisNec-style panel and reaches comparable aggregate outcomes to Shortcut Score and confidence. The panel has 67.52\% macro Jaccard overlap with Shortcut Score, 43.36\% with VisNec-style, and 64.74\% with confidence, showing that these outcomes arise from distinct selections. A direct probability-gap average yields 31.19 Visual Gain at 50\%, supporting model-wise rank normalization whenever likelihood scales are aggregated across architectures. Overall, the label-free VCA panel transfers its prediction-aligned audit semantics to unlabeled benchmark triage.

\section{Object-Swap Construction Details}
\label{sec:supp_swap_audit}

The constructor seeded-shuffled 440 positive relation groups; 114 passed fixed center-gap, overlap, and image-boundary checks, while 326 failed the in-bounds swap constraint. All 114 feasible edits entered the construct audit. The renderer requires a post-swap center gap of at least 0.05 image extent and new-box IoU at most 0.20, removes each source patch with 4-pixel padding and Telea inpainting radius 3, and pastes with a 3-pixel feather. These values are fixed before model scoring. Three distinct non-author volunteers received no compensation. After brief instruction on the label definitions and workflow, all gave informed consent. Institutional policy did not require formal ethics review for this low-risk annotation task. Each independently assessed a separately shuffled, blinded batch. Model margins, outcome strata, and sample identifiers were hidden, and 10\% repeated items measured within-annotator consistency. Majority consensus required the original relation to be valid, the edited relation to flip, both edited objects to remain visible, and artifacts not to block judgment. Consensus accepted 108 edits, rejected four, and left two three-way ties unresolved. Pairwise agreement across the four fields is 90.12--96.30\%, and repeat agreement is 100\%. The low-risk task involved judgments of edited public-dataset images; no demographic, behavioral, or sensitive personal data were collected.

\par\noindent\begin{minipage}{\columnwidth}
\centering
{\small
\setlength{\tabcolsep}{3pt}
\begin{tabular*}{\columnwidth}{@{\extracolsep{\fill}}L{0.64\columnwidth}rr@{}}
\toprule
Audit outcome & \(N\) & Rate \\
\midrule
Geometry-feasible candidates & 114 & 100.00 \\
Accepted by majority consensus & 108 & 94.74 \\
Rejected by majority consensus & 4 & 3.51 \\
No majority (three-way tie) & 2 & 1.75 \\
\bottomrule
\end{tabular*}}
\captionof{table}{Object-swap construct audit after majority consensus.}
\label{tab:supp_swap_audit}
\end{minipage}\par\medskip

The audit reduces the concern that the reported flip target is driven by unreadable edits. It does not turn patch swaps into photorealistic semantic edits; the no-edit SRC analysis provides a complementary validation without that rendering concern.

\paragraph{Feasibility-selection audit.} The 114/440 candidate rule is fixed geometry: no model outputs, VCA scores, or edited outcomes enter selection. We compare these candidates with the 326 excluded source pairs using pair bootstrap intervals.

\par\noindent\begin{minipage}{\columnwidth}
\centering
{\scriptsize
\setlength{\tabcolsep}{1.2pt}
\begin{tabular*}{\columnwidth}{@{\extracolsep{\fill}}lrrr@{}}
\toprule
Source characteristic & Feasible & Excluded & Difference [95\% CI] \\
\midrule
Horizontal relation & 69.30 & 61.35 & $+7.95$ [$-2.18$,$+17.99$] \\
Larger-box area & 13.91 & 23.18 & $-9.27$ [$-11.01$,$-7.55$] \\
Center gap & 42.68 & 50.38 & $-7.70$ [$-10.10$,$-5.31$] \\
Destination slack & 5.30 & $-10.99$ & $+16.29$ [$+15.15$,$+17.47$] \\
Qwen source D-CC & 91.23 & 83.74 & $+7.49$ [$+0.60$,$+13.98$] \\
InternVL source D-CC & 59.65 & 46.63 & $+13.02$ [$+2.55$,$+23.63$] \\
LLaVA source D-CC & 88.60 & 85.58 & $+3.01$ [$-4.40$,$+9.90$] \\
Ministral source D-CC & 65.79 & 62.27 & $+3.52$ [$-6.79$,$+13.61$] \\
\bottomrule
\end{tabular*}}
\captionof{table}{Geometry-feasibility selection audit over all 440 source pairs. Areas, distances, rates, and D-CC are percentages of image extent or examples as applicable.}
\label{tab:supp_swap_selection}
\end{minipage}\par\medskip

Feasibility favors smaller boxes with sufficient destination clearance; the horizontal/vertical composition difference is not distinguishable from zero. Qwen and InternVL source outcomes are easier on the feasible subset, whereas LLaVA and Ministral intervals cross zero. The audit excludes outcome-driven selection but limits generalization of the edit experiment to geometry-compatible object swaps.

\section{Non-Core Probe Sensitivity}

Table~\ref{tab:supp_probe_sensitivity} separates the primary Text+Blank definition from two stricter diagnostics. Strict T/B requires both controls to favor the opposite original answer. Dep+M adds random mismatch to the maximum-support control, while Strict+M combines mismatch with opposite-answer isolation. These columns are not alternative headline scores.

\par\noindent\begin{minipage}{\columnwidth}
\centering
{\scriptsize
\setlength{\tabcolsep}{1.8pt}
\begin{tabular*}{\columnwidth}{@{\extracolsep{\fill}}llrrrrr@{}}
\toprule
Model & Data & Dep & Strict T/B & Dep+M & Strict+M & Mean \(G^{\mathrm{img}}\) \\
\midrule
Qwen & VSR & 71.10 & 48.68 & 62.21 & 47.72 & 2.476 \\
Qwen & GSR-COCO & 81.82 & 42.84 & 74.89 & 40.23 & 2.953 \\
InternVL & VSR & 57.01 & 6.57 & 50.52 & 3.36 & 0.297 \\
InternVL & GSR-COCO & 42.84 & 10.45 & 37.73 & 8.07 & -0.264 \\
LLaVA & VSR & 76.38 & 21.54 & 75.58 & 20.58 & 1.072 \\
LLaVA & GSR-COCO & 64.55 & 5.91 & 59.66 & 5.45 & 0.651 \\
Ministral & VSR & 63.33 & 13.93 & 52.12 & 5.44 & 0.898 \\
Ministral & GSR-COCO & 59.77 & 24.43 & 45.00 & 6.82 & 0.605 \\
\bottomrule
\end{tabular*}}
\captionof{table}{Sensitivity to stricter and mismatch-inclusive definitions.}
\label{tab:supp_probe_sensitivity}
\end{minipage}\par\medskip

Mismatch lowers Dep by 0.80--14.77 points across these settings, but it does not change the qualitative finding that accuracy exceeds D-CC. The much larger collapse under strict isolation is explained by its answer-direction requirement rather than by a lack of image contribution alone.

\subsection{Matched Image-Permutation Calibration}

The aggregate calibration preserves the exact audited decision: \(d_i=\mathrm{sign}(m_{i,o})\) remains fixed while the original-image margin is replaced by the cached deterministic same-split image-permutation margin. This asks whether unrelated visual content reproduces support for the original decision under the same text-only and blank controls.

\par\noindent\begin{minipage}{\columnwidth}
\centering
{\scriptsize
\setlength{\tabcolsep}{1.4pt}
\begin{tabular*}{\columnwidth}{@{\extracolsep{\fill}}llrrr@{}}
\toprule
Model & Data & D-CC & Perm. D-CC & \(\Delta_{\rm DCC}^{\rm perm}\) [95\% CI] \\
\midrule
Qwen & VSR & 58.53 & 31.22 & 27.30 [23.38,31.31] \\
Qwen & GSR-COCO & 78.18 & 39.32 & 38.86 [35.23,42.50] \\
InternVL & VSR & 39.55 & 13.53 & 26.02 [23.46,28.74] \\
InternVL & GSR-COCO & 34.43 & 13.18 & 21.25 [18.64,23.86] \\
LLaVA & VSR & 51.56 & 3.76 & 47.80 [44.92,50.76] \\
LLaVA & GSR-COCO & 55.68 & 13.07 & 42.61 [39.77,45.34] \\
Ministral & VSR & 48.52 & 14.09 & 34.43 [31.63,37.23] \\
Ministral & GSR-COCO & 55.34 & 22.05 & 33.30 [30.57,36.02] \\
\bottomrule
\end{tabular*}}
\captionof{table}{Fixed-direction image-permutation calibration (\%). \(\Delta_{\rm DCC}^{\rm perm}\) is the paired difference between original-image and permutation D-CC while the original decision and correctness label remain fixed. Intervals use 5,000 paired bootstrap resamples, clustered by GSR relation pair.}
\label{tab:supp_null_calibration}
\end{minipage}\par\medskip

The original image yields 21.25--47.80 more D-CC points in all eight runs, and every paired interval excludes zero. Because accuracy is fixed, \(\mathrm{C\mbox{-}U}^{\rm perm}-\mathrm{C\mbox{-}U}^{\rm orig} =\Delta_{\rm DCC}^{\rm perm}\) exactly. Correct original-only assignments span 22.73--49.72\% of each run, whereas correct permutation-only assignments span 0.91--15.13\%. Thus the result is a paired calibration of the headline correctness--credit decomposition, not only of the all-decision Dep rate.

\subsection{Control Aggregation}

The max-control definition is fixed by the conjunction in VCA: positive credit requires original-image support to exceed each no-image control. We nevertheless compare signed alternatives and distributional distances using the same cached margins. Agreement and Spearman $\rho$ are summarized across all eight model--dataset runs. No variant is tuned on either external target.

\par\noindent\begin{minipage}{\columnwidth}
\centering
{\scriptsize
\setlength{\tabcolsep}{1.5pt}
\begin{tabular*}{\columnwidth}{@{\extracolsep{\fill}}lrrrrr@{}}
\toprule
Signal & Mean agr. & Min agr. & $\rho$ & SRC & Swap \\
\midrule
$r_o-r_t$ & 88.85 & 71.59 & 0.899 & 71.28 & 66.20 \\
$r_o-r_b$ & 90.83 & 81.14 & 0.920 & 71.41 & 69.03 \\
$r_o-(r_t+r_b)/2$ & 88.92 & 77.61 & 0.979 & 72.59 & 68.63 \\
$r_o-\max(r_t,r_b)$ & 100.00 & 100.00 & 1.000 & 72.97 & 69.15 \\
Probability max & 100.00 & 100.00 & 0.947 & 68.22 & 63.38 \\
Minimum JS & 64.97 & 42.84 & 0.595 & 66.76 & 64.34 \\
\bottomrule
\end{tabular*}}
\captionof{table}{Control aggregation sensitivity across eight runs. External columns macro-average within-model AUROC; Swap is the primary flip+visibility target on the 108 accepted edits.}
\label{tab:supp_control_aggregation}
\end{minipage}\par\medskip

The signed margin variants retain strong agreement and external ranking, so the result is not peculiar to one score parameterization. Probability differences preserve every binary credit decision but compress within-model rankings. Exponentiating a signed log-odds gap gives an odds ratio with identical ordering. Unsigned JS instead detects any distribution change, including movement away from the original decision, and is therefore not a substitute for prediction-aligned credit. We do not fit a label-calibrated probability variant, because doing so would introduce the held-out labels that the dependence event is designed to avoid. On the independent Swap target, max-control AUROC is 69.15 versus 66.20, 69.03, and 68.63 for text-only, blank-only, and mean aggregation, respectively; original confidence reaches 67.90. These are unadjusted rankings, while the edit-grouped incremental analysis above remains the adjusted test.

\par\noindent\begin{minipage}{\columnwidth}
\centering
{\scriptsize
\setlength{\tabcolsep}{2.0pt}
\begin{tabular*}{\columnwidth}{@{\extracolsep{\fill}}lrrrr@{}}
\toprule
Control set & Credit & C-U & Agr. & C-U Jacc. \\
\midrule
Text + fixed blank & 65.01 & 19.07 & 100.00 & 100.00 \\
Text + mean-color blank & 62.55 & 20.89 & 96.44 & 88.01 \\
Text + fixed blank + mismatch & 57.71 & 25.02 & 92.71 & 76.21 \\
\bottomrule
\end{tabular*}}
\captionof{table}{Micro-pooled item stability over 8,516 records. C-U is correct-but-uncredited as a fraction of all records; agreement and C-U Jaccard are relative to the headline fixed-blank definition.}
\label{tab:supp_control_set_stability}
\end{minipage}\par\medskip

Across all three sets, 7,646/8,516 (89.78\%) records retain the same credit assignment. Their C-U intersection contains 1,593 records and their union 2,274 (18.71\% and 26.70\% of all records); 1,593/1,624 (98.09\%) headline C-U records remain in the intersection. Per-run stable assignment spans 78.46--98.48\%, and headline C-U retention spans 90.95--100.00\%. Thus added controls enlarge the boundary set but rarely reverse the original C-U core.

\subsection{Cached-Context Proxy Comparison}

We adapt named modality-use and grounding metrics to the same cached records. Perceptual Score is the sample accuracy difference between the original and the fixed same-split image permutation, and therefore uses benchmark labels. VisNec-style is the gold-answer loss reduction from the cached text-only control to the original image; unlike exact VisNec, it does not use model-specific visual-token masking or semantic clustering. Permutation drop is the prediction-aligned probability decrease under that permutation. For the pixel-adapted FPVG and sufficiency/comprehensiveness baselines, IrrE approximates a relevant-only view and RelE an irrelevant-only view. This adaptation is not the original detector-feature implementation, and inherits the erasure limitations detailed below. No baseline is tuned on either external target.

Fu et al. (2026) define a label-dependent per-instance null-input baseline and modality gain, contrast images with generated captions, and report normalized Visual Dependence at the model--benchmark level. VCA instead assigns a label-free event and signed gap to each fixed decision and separately tests same-image relation interaction. For a direct common-record comparison, we adapt their normalized accuracy-drop view to the mean accuracy of our text-only and blank controls:

\par\noindent\begin{minipage}{\columnwidth}
\centering
{\small
\setlength{\tabcolsep}{3pt}
\begin{tabular*}{\columnwidth}{@{\extracolsep{\fill}}lrr@{}}
\toprule
Model & VSR aggregate VD & GSR-COCO aggregate VD \\
\midrule
Qwen & 42.67 & 45.94 \\
InternVL & 9.21 & 13.72 \\
LLaVA & 29.78 & 38.83 \\
Ministral & 17.64 & 30.29 \\
\bottomrule
\end{tabular*}}
\captionof{table}{Label-dependent aggregate visual dependence, using the normalized accuracy drop from original to mean text-only/blank accuracy.}
\label{tab:supp_aggregate_vd}
\end{minipage}\par\medskip

\par\noindent\begin{minipage}{\columnwidth}
\centering
{\scriptsize
\setlength{\tabcolsep}{1.5pt}
\begin{tabular*}{\columnwidth}{@{\extracolsep{\fill}}lrrrr@{}}
\toprule
& \multicolumn{2}{c}{SRC (4 models)} & \multicolumn{2}{c}{Swap strict (4 models)} \\
Signal & AUROC & AUPRC & AUROC & AUPRC \\
\midrule
Confidence & \textbf{73.66} & \textbf{69.73} & \textbf{76.45} & \textbf{63.73} \\
VisNec-style & 58.52 & 56.83 & 50.69 & 38.50 \\
Perceptual Score & 62.23 & 57.32 & 53.50 & 40.52 \\
Permutation drop & 64.31 & 60.65 & 60.72 & 48.75 \\
FPVG-style & 59.89 & 56.28 & 51.37 & 39.32 \\
$-$Sufficiency & 51.10 & 51.67 & 46.65 & 34.74 \\
Comprehensiveness & 63.78 & 60.05 & 60.31 & 47.04 \\
Comp.--Suff. & 64.13 & 60.11 & 61.71 & 47.53 \\
Raw image gap & 72.97 & 67.09 & 75.39 & 61.25 \\
\bottomrule
\end{tabular*}}
\captionof{table}{Macro-average within-model cached-context proxy comparison. Swap uses all four models on the same 108 accepted edits and the strict composite target; named baselines use common-context adaptations.}
\label{tab:supp_prior_metrics}
\end{minipage}\par\medskip

Under these cached-context adaptations, confidence and the VCA gap are the two strongest signals on both checks; confidence is slightly stronger. Among the adapted modality-ablation and region-grounding signals, the VCA gap has the strongest macro AUROC. These common-context adaptations isolate score behavior on identical records; the original methods' model-specific masking, auxiliary text model, and semantic clustering remain outside this comparison.

\subsection{Surface-Form Robustness}

We score the complete VSR and GSR-COCO test sets under the canonical Yes/No interface, a semantically matched True/False interface, and an untuned ensemble that averages their aligned margins in each context. The ensemble uses no labels, fitting, or development selection.

\par\noindent\begin{minipage}{\columnwidth}
\centering
{\scriptsize
\setlength{\tabcolsep}{0.8pt}
\begin{tabular*}{\columnwidth}{@{\extracolsep{\fill}}llrrrrrr@{}}
\toprule
& & \multicolumn{3}{c}{True/False} & \multicolumn{3}{c}{Ensemble} \\
Model & Data & Acc & Dep & D-CC & Acc & Dep & D-CC \\
\midrule
Qwen & VSR & 76.06 & 61.65 & 54.92 & 80.22 & 67.17 & 58.53 \\
Qwen & GSR & 84.43 & 74.77 & 72.73 & 89.66 & 80.45 & 77.95 \\
InternVL & VSR & 62.93 & 66.69 & 47.40 & 64.85 & 67.57 & 48.68 \\
InternVL & GSR & 67.50 & 74.55 & 55.68 & 76.25 & 64.09 & 54.09 \\
LLaVA & VSR & 77.82 & 70.30 & 53.64 & 74.86 & 72.78 & 51.40 \\
LLaVA & GSR & 81.14 & 75.57 & 67.16 & 83.07 & 67.05 & 58.98 \\
Ministral & VSR & 73.34 & 83.75 & 64.21 & 68.21 & 72.86 & 56.93 \\
Ministral & GSR & 83.41 & 90.68 & 78.52 & 79.77 & 76.25 & 69.43 \\
\bottomrule
\end{tabular*}}
\captionof{table}{Full-test multi-verbalizer results. Ensemble averages Yes/No and True/False margins before applying the unchanged VCA rule.}
\label{tab:supp_surface_robustness}
\end{minipage}\par\medskip

Under the ensemble, 10.34--24.09\% of decisions are correct but uncredited. Against canonical Yes/No, ensemble agreement spans 73.75--95.11\% for original predictions, 63.30--92.55\% for Dep, and 72.39--97.44\% for D-CC. Accuracy rankings are unchanged on VSR and GSR-COCO ($\rho=1.00$); D-CC rankings remain strongly aligned ($\rho=0.80$ on each). These complete-test results support the prespecified ensemble as an aggregate surface check. Where inverse relation statements are available, the pair-centered contrast below more directly cancels answer-surface offsets shared by the true and false statements.

\par\noindent\begin{minipage}{\columnwidth}
\centering
{\scriptsize
\setlength{\tabcolsep}{1.0pt}
\begin{tabular*}{\columnwidth}{@{\extracolsep{\fill}}llrrrr@{}}
\toprule
Model & Data & \(n\) & Item flip & Pair flip & Pair Jacc. \\
\midrule
Qwen & VSR & 504 & 23.61 & 3.37 & 96.23 \\
Qwen & GSR & 440 & 16.82 & 2.27 & 97.60 \\
InternVL & VSR & 504 & 47.22 & 27.98 & 62.80 \\
InternVL & GSR & 440 & 49.89 & 20.91 & 72.70 \\
LLaVA & VSR & 504 & 8.73 & 5.36 & 93.95 \\
LLaVA & GSR & 440 & 29.89 & 3.64 & 96.01 \\
Ministral & VSR & 504 & 26.39 & 10.91 & 86.59 \\
Ministral & GSR & 440 & 34.09 & 5.23 & 94.21 \\
\bottomrule
\end{tabular*}}
\captionof{table}{Yes/No--True/False event stability on identical records. Item flip is per-question dependence disagreement; Pair flip and Jacc. use \(\mathrm{PairCredit}={\bf 1}\{G^{\rm int}>0\}\).}
\label{tab:supp_pair_verbalizer}
\end{minipage}\par\medskip

GSR uses all 440 benchmark-provided true/false groups. For VSR, we construct inverse queries only for positive items with structured subject and object fields in the six strict families left/right, above/below, and front/behind. This yields 504 same-image, same-object-pair groups; 35 candidates without structured entities are excluded before scoring. Pair differencing lowers event flips in every model--dataset comparison, although the InternVL/VSR residual shows that verbalizer--prompt interaction is not purely an additive token offset.

We separately use the same 200 VSR examples, stratified over above, below, left, and right, for diagnostics not covered by the full test: three relation wordings and a neutral Qwen system instruction. Paraphrase rows pool 600 query instances; agreement averages the two synonyms against the canonical wording.

\par\noindent\begin{minipage}{\columnwidth}
\centering
{\scriptsize
\setlength{\tabcolsep}{1.5pt}
\begin{tabular*}{\columnwidth}{@{\extracolsep{\fill}}llrrrr@{}}
\toprule
Form & Model & Acc & D-CC & Pred. & Credit \\
\midrule
Paraphrase & Qwen & 78.00 & 64.00 & 83.75 & 79.25 \\
Paraphrase & InternVL & 59.67 & 42.17 & 97.00 & 74.25 \\
Paraphrase & LLaVA & 67.00 & 54.67 & 84.25 & 79.00 \\
System & Qwen & 81.00 & 65.50 & 88.00 & 82.00 \\
\bottomrule
\end{tabular*}}
\captionof{table}{Additional fixed-200 VSR wording diagnostics.}
\label{tab:supp_paraphrase_robustness}
\end{minipage}\par\medskip

The selected yes, no, true, and false continuations are one token under each model's scoring backend, so the comparison does not mix answer identity with continuation length. We do not treat unequal multi-token sentences as equivalent logits. The Qwen system diagnostic uses the neutral instruction ``Answer the user's question with exactly yes or no,'' rather than prompting the model to rely more heavily on the image.

\subsection{Visual-Control Robustness}

We replace the fixed mid-gray canvas with a per-image uniform mean color on the complete VSR and GSR-COCO tests for all four models. Both controls are content-free; mean color additionally preserves each image's dimensions and first-order color average. Original and text-only scores are unchanged.

\par\noindent\begin{minipage}{\columnwidth}
\centering
{\scriptsize
\setlength{\tabcolsep}{1.1pt}
\begin{tabular*}{\columnwidth}{@{\extracolsep{\fill}}llrrrrrr@{}}
\toprule
Model & Data & $N$ & Acc & Fixed & Mean & Agree & $\rho$ \\
\midrule
Qwen & VSR & 1249 & 79.42 & 58.53 & 57.65 & 98.00 & .997 \\
Qwen & GSR & 880 & 90.91 & 78.18 & 77.61 & 99.09 & .996 \\
InternVL & VSR & 1249 & 58.69 & 39.55 & 38.75 & 97.92 & .993 \\
InternVL & GSR & 880 & 58.41 & 34.43 & 33.75 & 98.52 & .996 \\
LLaVA & VSR & 1249 & 69.90 & 51.56 & 50.92 & 99.04 & .999 \\
LLaVA & GSR & 880 & 81.93 & 55.68 & 51.93 & 95.91 & .992 \\
Ministral & VSR & 1249 & 64.45 & 48.52 & 44.76 & 89.43 & .912 \\
Ministral & GSR & 880 & 71.48 & 55.34 & 51.36 & 94.20 & .962 \\
\bottomrule
\end{tabular*}}
\captionof{table}{Full-test fixed-gray versus mean-color controls. Fixed and Mean report D-CC; Agree is per-item credit agreement, and $\rho$ is Spearman correlation between continuous dependence gaps.}
\label{tab:supp_visual_controls}
\end{minipage}\par\medskip

Mean color changes D-CC by $-0.57$ to $-3.98$ points, with 89.43--99.09\% credit agreement and $\rho=0.912$--0.999. Under this alternate control, D-CC remains 13.30--30.00 points below accuracy in every model--dataset run. The central empirical gap is therefore not specific to a single gray value, while the shifts quantify residual control sensitivity. A separate fixed-200 pilot also included strong blur and deterministic $8\!\times\!8$ patch shuffle. Because these retain or degrade semantic content, they are stress probes rather than empty controls and do not enter the headline credit definition.

\section{Reproducibility Details}

Each run contains matched original, text-only, and blank margins. We group by sample id, fix prediction direction, and preserve 440 supplied GSR-COCO pairs; no stochastic sampling, learned scorer, or label-selected threshold is used. Dependence intervals use 5,000 bootstraps (seed 20260719), and semantic H--L intervals use 2,000. GSR intervals resample relation-pair ids, and swap intervals resample edit ids while retaining all model records in each cluster. Factorial model-level conditional proportions use two-sided 95\% Wilson intervals. Pooled intervals use 5,000 resamples of procedural stem ids with seed 20260724 and retain all four model records for each sampled stem. Generation is fixed before model scoring and checked for complete cells, direction balance, and exact context reuse. Cross-backbone summaries macro-average metrics computed within model. Qwen, InternVL, and LLaVA scoring uses Python 3.13.5, PyTorch 2.6.0+cu124, and Transformers 4.57.3; Ministral uses Python 3.13.5, the same PyTorch build, and Transformers 5.15.0.dev0 for model support. Experiments run on eight NVIDIA RTX A6000 48GB GPUs with native model processors and bfloat16 model weights (Ministral resolves its checkpoint-native automatic dtype). Prompt serialization, selected answer strings, continuation losses, model predictions, and margins are stored per record rather than reconstructed from aggregate tables. The anonymous submission artifact includes the complete source, exact prompt and chat serialization, deterministic GSR conversion and factorial builders, all 1,800 factorial inputs per model, accepted edits, and deidentified annotation decisions.

\end{document}